\documentclass[lettersize,journal]{IEEEtran}
\usepackage{amsmath,amsfonts}
\usepackage{algorithmic}
\usepackage{algorithm}
\usepackage{array}
\usepackage[caption=false,font=normalsize,labelfont=sf,textfont=sf]{subfig}
\usepackage{textcomp}
\usepackage{stfloats}
\usepackage{url}
\usepackage{verbatim}
\usepackage{graphicx}
\usepackage{cite}

\usepackage{amsmath}
\usepackage{amssymb}
\usepackage{mathtools}
\usepackage{amsthm}
\usepackage{booktabs}

\usepackage{array}
\usepackage{multirow}
\usepackage{arydshln}
\usepackage{makecell}
\usepackage{amsthm}
\newcommand{\method}[2]{\hyperlink{cite.#2}{#1}}
\usepackage{lipsum} 
\usepackage{hyperref}

\newtheorem{theorem}{\textbf{Theorem}} 
\newtheorem{remark}{Remark} 
\newtheorem{lemma}{Lemma} 

\newtheorem{definition}{Definition}
\newtheorem{researchq}{RQ}

\begin{document}

\title{Unsupervised Multi-modal Feature Alignment \\ for Time Series Representation Learning}

\author{Chen Liang, Donghua Yang, Zhiyu Liang, Zheng Liang, Xiyang Zhang, Hongzhi Wang, Jianfeng Huang, \\
Email: 23B903050@stu.hit.edu.cn,  yang.dh, zyliang, lz20@hit.edu.cn, 7203610125@stu.hit.edu.cn, wangzh@hit.edu.cn, JFHuang.research@gmail.com     
\thanks{Hongzhi Wang is the corresponding author.}}

\markboth{Journal of \LaTeX\ Class Files,~Vol.~14, No.~8, August~2021}%
{Shell \MakeLowercase{\textit{et al.}}: A Sample Article Using IEEEtran.cls for IEEE Journals}

\IEEEpubid{0000--0000/00\$00.00~\copyright~2021 IEEE}

\maketitle

\begin{abstract}
  Unsupervised representation learning (URL) for time series data has garnered significant interest due to its remarkable adaptability across diverse downstream applications. It is challenging to ensure the utility for downstream tasks by only identifying patterns from the temporal domain features since the goal of URL differs from supervised learning methods. A variety of time series transform techniques, e.g., spectral features, wavelet transformed features, features in image form and symbolic features, etc., transform time series to multi-modal informative features. This study proposes an alignment-based method to harvest transform properties from them. In contrast to common methods that fuse features from multiple modalities, our proposed method simplifies the final neural architecture by retaining a single time series encoder. By validating the proposed method on a diverse set of time series datasets, our approach outperforms existing state-of-the-art URL methods across 3 diverse downstream tasks. This paper introduces a novel model-agnostic paradigm for time series URL, paving a new research direction. Code available at \url{https://anonymous.4open.science/r/MMFA-8E8B}.
\end{abstract}


\section{Introduction}
Time series describes a variable over time encompassing monitoring measures. Multivariate time series (MTS) involves sets of dependent variables, playing a pivotal role across domains from finance to healthcare and the natural sciences~\cite{bennett2022detection, kim2019fault,riley2021internet, bi2023accurate}.

\textbf{Unsupervised representation learning for time series.} Time series pose complexities in pattern identification and analysis. This further increases the difficulty of obtaining labels for supervised learning~\cite{tonekaboni2021unsupervised}. Thus, unsupervised representation learning (URL) for time series~\cite{eldele2021time, yang2022unsupervised, liang2023contrastive,yue2022ts2vec} has emerged. \textit{URL aims to train a neural network (called encoder), without requiring labels, to encode the data into geometric signatures (or embedding), to capture inherent patterns of the raw data.} The learned representation, necessitating minimal annotation~\cite{zerveas2021transformer}, proves valuable in a variety of downstream tasks including temporal data classification, clustering~\cite{paparrizos2023odyssey, bonifati2022time2feat}, anomaly detection~\cite{liang2023contrastive, yue2022ts2vec, schmidl2022anomaly}, retrieval~\cite{zhu2020deep, paparrizos2019grail}, etc. 

\textbf{Representation arrangement mechanisms for URL.} Moreover, akin to deep clustering~\cite{guo2017improved} and manifold learning~\cite{balasubramanian2002isomap} methods, URL can help the encoders better capture local~\cite{pan2014evaluation,he2003locality} or global~\cite{batista2011complexity} structures, which can be induced from patterns present in the original samples and learn from prior deductive knowledge. As a result, unsupervised methods with simple classification heads on learned representations may outperform methods tailored for downstream tasks~\cite{liang2023contrastive}.

\textbf{Transforms as feature engineering for time series.} Researchers have explored transforming time series data into image~\cite{wang2015encoding,park2023meta} or symbol bags~\cite{schafer2012sfa, tang2020interpretable} to comprehend them better. Various feature engineering methods are proposed to parse time series and expose abundant informative patterns that are more easily identified. However, the challenges for leveraging these transforms can be further explored as follows. 

\textbf{Challenge 1: Disjoint transform properties of multi-modal features} Invariance refers to representation unchanged when input changes, while equivariance refers to representation changing with input, according to certain rules~\cite{guto2023icml}. They are \textbf{transform properties} of the transforms and encoders. These transform properties can introduce consistent but disjoint prior knowledge. Learning from them individually introduces gaps between learned transform properties~\cite{yang2022unsupervised}. Besides, the encoder merely needs to identify easily learnable patterns during feature fusing~\cite{ren2022co}, without considering algorithmic associations between different multi-modal patterns, causing learning from shortcuts and blended blind source noises. This can lead to the learning of false patterns~\cite{robinson2021can}, resulting in information loss~\cite{le1988preservation}, data misinterpretation~\cite{geirhos2020shortcut}, and decreased model performance~\cite{puli2023don}.

\textbf{Challenge 2: Coupled transforms and encoders during inference.} Feature fusion structures may overly complicate the neural network~\cite{yang2022unsupervised,tang2020interpretable, park2023meta}. Most feature engineering methods for time series are highly coupled with the neural structure of the encoder. For example, a dual tower structure fuses time and spectral domain features, filling the inductive gap between these views. It is necessary to acquire Fourier-transformed features first and encode both features whenever we use the encoder. The cost would be unbearable if more time-consuming transforms and encoders like continuous wavelet transform~\cite{grossmann1990reading} and a 2D ResNet architecture~\cite{wang2015encoding} were involved during inference.

\textbf{Overcome the challenges.} To overcome the above challenges, we inject prior inductive bias implemented by multi-modal transforms into the raw time series encoder and propose a model-agnostic framework. This encoder recovers the most salient local and global arrangement preserved by various transforms and feature extraction processes. 
\newpage

\textbf{Methods.} We treat the encoders as a graph spectral embedding learner on the graph which aligns the multi-modal features. The alignment with the transforms fine-tunes the global arrangement of the representations, which originally preserves only the local arrangement of the raw time series.
Therefore, only preserving the neural encoder for the temporal domain is adequate during inference. Our analysis demonstrates that this framework significantly outperforms state-of-the-art methods on multiple downstream tasks.

The main contributions of this paper are summarized as follows:

\begin{itemize}
\item We introduce a novel feature-aligned URL framework, leveraging intricate feature engineering and signal processing methods, outperforming currently available methods while preserving high scalability.

\item The proposed method learns local and global arrangement of multi-modal features, and sufficiently leverages different inductive biases introduced.

\item The close-formed solutions of the proposed objectives presents recovery of classic manifold learning methods, pointing toward a new model-agnostic and theoretically-oriented URL for TS research direction.

\item Through experiments on time series datasets in different domains and with different characteristics (i.e., length, dimensionality, and train set sizes), our proposed approach outperforms all existing URL approaches by a large margin, better than approaches tailored to multiple downstream tasks.
\end{itemize}

\begin{figure}[t]
  \centerline{\includegraphics[width=\linewidth]{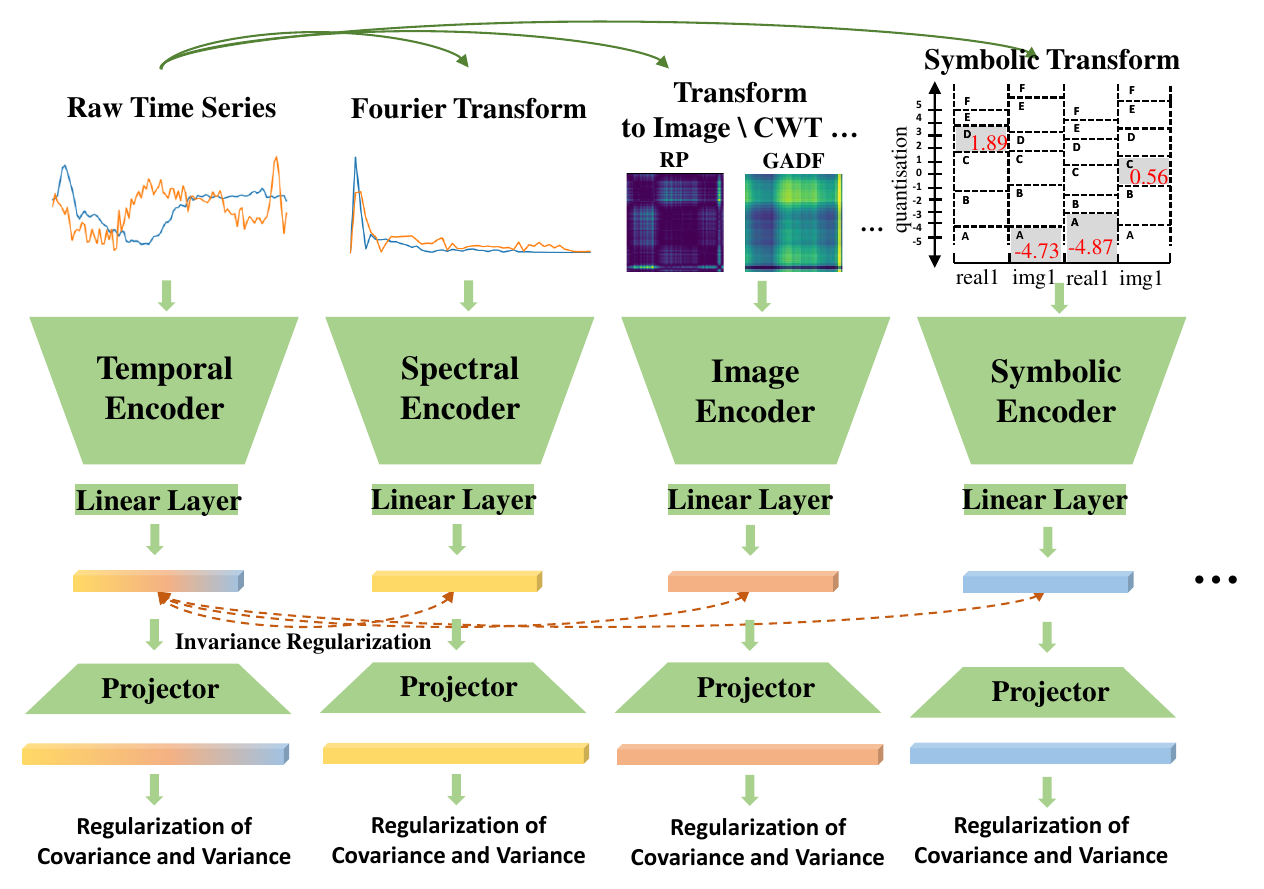}}
  \caption{\textbf{An overview of the \textbf{MMFA} framework.} A range of transforms is applied to raw time series to produce informative features across multiple modalities. These features are subsequently processed by neural feature extractors to identify different patterns. Following this, the representations are mapped to an embedding space and aligned via the regularization method.}
  \label{overview}
\end{figure}

\section{Related Work}\label{sec:rela}

In this section, we overview the research in the landscape related to the domain of our work. We categorize them into three themes and discuss their contributions, limitations, and relevance to our work.

\subsection{Representation Learning for Time Series}

\textbf{Introducing representation learning for time series} Unsupervised representation learning provides a scalable approach to obtain meaningful representations~\cite{meng2023unsupervised}. Neural architectures such as autoencoders and seq2seq models~\cite{vadiraja2020survey} have paved the way for methods based on data reconstruction~\cite{TiMAE, GeneralizableAR}, and context prediction to learn representations~\cite{ma2019learning, malhotra2017timenet}. Contrastive learning distinguishes the strict positive (augmented by the same sample) and negative (positive) sample pairs to capture invariance after augmentation, which is presented as an important method~\cite{huynh2022boosting,franceschi2019unsupervised, eldele2021time, yue2022ts2vec, wu2022timesnet}. Reconstruction and generative methods~\cite{zerveas2021transformer,li2023ti}, can also produce their hidden states as dense representations. Representation learning not only unifies different tasks but also provides scalable ways for time series data management~\cite{paparrizos2019grail}.

\textbf{Learning arrangements with different global and local levels.} Notably, URL encompasses multi-levels of learning objectives: instance level~\cite{chen2020big, oord2018representation,chen2020simple}, cluster (prototype) level~\cite{li2020prototypical, caron2020unsupervised, meng2023mhccl}, and temporal level~\cite{tonekaboni2021unsupervised, eldele2021time, hyvarinen2016unsupervised}, etc., each capturing different aspects of similarity and dependency in data, representing preservation of \textit{local} or \textit{global} arrangement of the temporal data.

On time series, instance-level contrastive learning perturbs original samples, emphasizing the retention of salient features~\cite{yue2022ts2vec}. In this case, the proposed supervision signal enhances the preservation of the \textit{local} arrangement of data sampled from a distribution of permuted samples. It works well when downstream task goals are sensitive to \textit{local} fluctuations. 

Cluster-level contrastive learning extends the focus to the \textit{global} shape of sample distributions~\cite{meng2023mhccl}. Meanwhile, temporal-level contrastive learning captures temporal dependencies surrounding each sample, learning the causal structure of time series data as \textit{global} arrangement~\cite{eldele2021time}. GRAIL~\cite{paparrizos2019grail} learns temporal representation vectors to reconstruct the distance matrix with the temporal invariance distance measure. These methods \textit{globally} align \textit{local} structures captured naturally by the encoder, thus, injecting different global information into the feature encoder.

\subsection{Feature Transforms}\label{sec:transform_main}

Researchers have investigated various signal processing and feature engineering techniques. These techniques help preserve salient patterns that are challenging to capture without additional prior knowledge. (1). \textbf{Discrete Fourier transform (DFT)}~\cite{yang2022unsupervised,winograd1978computing}, which reveals the spectral structure of the original time series. In \cite{yang2022unsupervised}, the authors find that with the same training methods, URL using only DFT features or raw data share a small proportion of false prediction during evaluation, with the rest of them being non-overlapping, revealing relatively complementary inductive bias.  (2). \textbf{Continuous wavelet transform (CWT)} CWT is a tool that provides an overcomplete representation of a signal by letting the translation and scale parameter of wavelets vary continuously~\cite{grossmann1990reading}. Compared to raw time series that collapse spectral features, or DFT obscuring temporal stages and trends, Continuous Wavelet Transform (CWT) arranges temporal and spectral patterns in a 2D plane. (3). \textbf{Image-like features} (i.e., GAF, RP~\cite{wang2015encoding}), which can be encoded leveraging the local precentral view of CNNs. \cite{wang2015encoding} proposes a framework to encode time series data as different types of images thus allowing machines to visually recognize and classify time series. E.g., using a polar coordinate system, Gramian Angular Field (GAF) images a represented as a Gramian matrix where each element is the trigonometric sum between different time intervals. (4). \textbf{Symbolic features} Techniques like Symbolic Aggregate Approximation (SAX)\cite{notaristefano2013data} or Symbolic Fourier Approximation (SFA)\cite{schafer2012sfa} transform noisy time series data into abstract symbolic patterns, filtering out noise and reducing classifier overfitting. SFA, a promising method for time series classification~\cite{tang2020interpretable}, aids in handling high-dimensional, sparse data prone to overfitting.

Generally, these methods provide overcomplete representations that are challenging to fully take advantage of in various tasks. These methods are always can be accelerated with GPU based parallel computation~\cite{ptwt, gupta2002scalability}.

Our algorithm learns global and local sample arrangements effectively, resulting in stronger theoretical guarantees and outperforming state-of-the-art URL methods.

\section{Preliminaries}

This section defines the key concepts used in the paper. First, we clarify the data type we study in this paper, multivariate time series. Then we introduce the transforms and encoders with their domain and codomain.

\begin{table}
  \centering
  \caption{Composition of Transform Operators and Neural Feature Extractors.}\label{tab:TFpairs}
  \begin{tabular}{p{0.11\textwidth}c:p{0.15\textwidth}cp{0.8\textwidth}cp{0.18\textwidth}}
      \toprule
      \multicolumn{2}{l:} {\textbf{Transform Operator}}& \multicolumn{2}{c} {\textbf{Feature Extractor}} \\
      \midrule
      \multicolumn{2}{l:} {Raw Time Series} & \multicolumn{2}{c}{ST} \\
      \hdashline
      \multicolumn{2}{l:}{FFT} & \multicolumn{2}{c}{ResNet1d} \\
      \hdashline
      \multirow{2}{*}{Image Encoding} & PR & \multicolumn{2}{c}{\multirow{5}{*}{ResNet12}}  \\
      \cdashline{2-2}
      & GADF \\
      \cdashline{1-2}
      \multirow{3}{*}{CWT} & db1   \\ 
      \cdashline{2-2}
      & dmey  \\ 
      \cdashline{2-2}
      & coif5   \\
      \cdashline{1-4}
      \multicolumn{2}{l}{\multirow{2}{*}{SFA}} & \multirow{2}{*}{Longformer} & Pre-trained \\
      \cdashline{4-4}
      & & & Random \\
      \bottomrule
  \end{tabular}
\end{table}

\subsection{Unsupervised Representation Learning for Multivariate Time Series} 

Multivariate time series is a set of variables, containing observations ordered by successive time. We denote a MTS sample with $D$ variables (a.k.a. dimensions or channels) and $T$ timestamps (a.k.a. length) as $x \in \mathbb{R}^{D \times T}$, and a dataset containing $N$ samples as the $X = \{x_1, x_2, \dots ,x_N\}\in\mathbb{R}^{N\times D \times T}$. 

Given an MTS dataset $X$, the goal of unsupervised representation learning (URL) is to train a neural network model (encoder) $f:\mathbb{R}^{D\times T} \to \mathbb{R}^{d_z}$, where $d_z$ is the dimensionality of the representation, the acquired representation $z_i = f(x_i)$ can be informative for downstream tasks, e.g., classification and forecasting. ``Unsupervised'' means that the labels of downstream tasks are unavailable when training $f$. To simplify the notation, we denote $f(X) = \{z_1, z_2, \dots, z_N\} \in \mathbb{R}^{N \times d_z}$ in following sections. 

\subsection{Multi-modal Features and Neural Encoders}\label{sec:neur_encoder_main}

Concerning raw MTS data and multi-modal features transformed, we denote the features as $x^{(i)} = T^{(i)}(x)$ in this work, and $i$ denotes the index of the selected transform. The transformed features are highly heterogeneous from the original raw MTS data. Therefore, we designed different neural architectures for the feature encoders w.r.t the multi-modal features, $z^{(i)} = f^{(i)}(x^{(i)})$. They can be composited as $f^{(i)} \circ T^{(i)}$. But the selection of $f^{(i)}$ and $T^{(i)}$ should be compatible. The compositions of them are listed in \textbf{Tab.}~\ref{tab:TFpairs}.

\subsection{Neural Encoder}\label{sec:neur_encoder}

In this section, we briefly introduce the encoders for all the modalities transformed from raw time series, and their domains and codomains, which are first mentioned in \textbf{Sec.}~\ref{sec:neur_encoder_main}.

\textbf{Shapelet Transformer (ST).} We use ST [1] as the backbone encoder after unsupervised learning. ST converts time series into energy sequences via learned wavelet-like shapelets, sliding across the signal and computing subsequence distances. By varying shapelet lengths (like wavelet dilation/compression), ST adaptively fits signal shapes and optimizes temporal-spectral resolution, akin to CWT. Here, ST serves as the raw temporal feature extractor \( f: \mathbb{R}^{D \times T} \to \mathbb{R}^{d_z} \), processing identity-transformed inputs \( T^{(id)}: \mathbb{R}^{D \times T} \to \mathbb{R}^{D \times T} \). Other encoders follow similarly.  

\textbf{CNN \& Transformer Feature Extractors.} For spectral and 2D patterns, we use ResNet CNNs—1D for DFT-transformed signals ($T^{(dft)}: \mathbb{R}^{D \times T} \to \mathbb{R}^{D \times T}$) and 2D for image-style ($T^{(img)}$) and CWT-based ($T^{(cwt)}$) representations, with interpolation ensuring fixed input dimensions. For symbolic features, we employ a transformer encoder: $T^{(sfa)}$ converts MTS into token sequences (length $L$) with embeddings fed into a pre-trained transformer, improving downstream performance via sequence modeling. Both architectures adapt to their respective input structures—CNNs for local patterns, transformers for sequential dependencies.

\textbf{Notations.} We denote \\ \( X^{(i)} = T^{(i)}(X)  = \{ x_1^{(i)}, x_2^{(i)}, \dots , x_N^{(i)} \} \in \mathbb{R}^{N \times \text{dim} T^{(i)}} \) \\ and  $ Z^{(i)} = f^{(i)}(X^{(i)}) =  \{ z^{(i)}_1, z^{(i)}_2, \dots , z^{(i)}_N\}, i \in [0,k]$ in subsequent sections, where $T^{(i)}$, $f^{(i)}$ are $i$th in $k$ predefined transform and encoder involved and $f^{(0)} = \text{Id}$ is an identity function. Eventually, we discuss the alignment graph and eigenfunctions defined on the sample space, which encompasses not only raw data but also transformed features. The domain of the eigenfunctions can be denoted as \( \mathcal{S}^{\text{\textit{data}}} = \sqcup_{i \in [0,k]} \text{Im}(T^{(i)}) \), given $k$ transforms, where $\text{Im}(T^{(i)})$ denotes \textbf{image/domains of the transforms} which is the space transformed features lies in.

\section{Overview}\label{sec:over}
In this section, we overview the framework of the proposed multi-modal feature alignment (MMFA). 

\textbf{Spliced images of transforms as entities to be represented.} Generally, we treat the transformed features as individual entities in the sample space $\mathcal{S}^{\text{\textit{data}}} = \sqcup_{i \in [0,k]} \text{Im}(T^{(i)})$. This procedure maps the euclidean space $\mathcal{S}^{\text{\textit{data}}}$ to an generic embedding space $\mathbb{R}^{d_z}$ through a piecewise function $\phi: \mathcal{S}^{\text{\textit{data}}} \to \mathbb{R}^{d_z}$. It can be defined as a piecewise function.

\begin{equation}
\begin{aligned}
        \forall i\in[0,k], \forall x \in \text{Im}(T^{(i)}) & \ \ \ \ \ \phi(x) = f^{(i)}(x)
\end{aligned}
\end{equation}

\textbf{Arragement of multi-modal entities.} For learning $\phi$ with high utility, leveraging spectral embedding theory, URL relies on an undirected graph $\mathcal{G}(\mathcal{X}, w)$ that indicates sample arrangement lies in the manifold in the original high dimensional space from where raw data is collected, where $\mathcal{X}\in [(k + 1)N]$ denotes the subscript for raw MTS and the transformed features. In this paper, the relation matrix $w \in \mathbb{R}^{(k + 1)N \times (k + 1)N}$ is constructed by connecting entities transformed from the same time series, and similar patterns lie in each of the feature modalities. 

\begin{equation}\label{eq:graph_weight}
w_{ij} =:   \begin{cases}  A_{T_{(\mathcal{X}_i)}, T_{(\mathcal{X}_j)}}(\mathcal{X}_i, \mathcal{X}_j) , i \neq j \\
 0, i = j 
\end{cases}
\end{equation}

\begin{figure*}
    \centering
    \includegraphics[width=\textwidth]{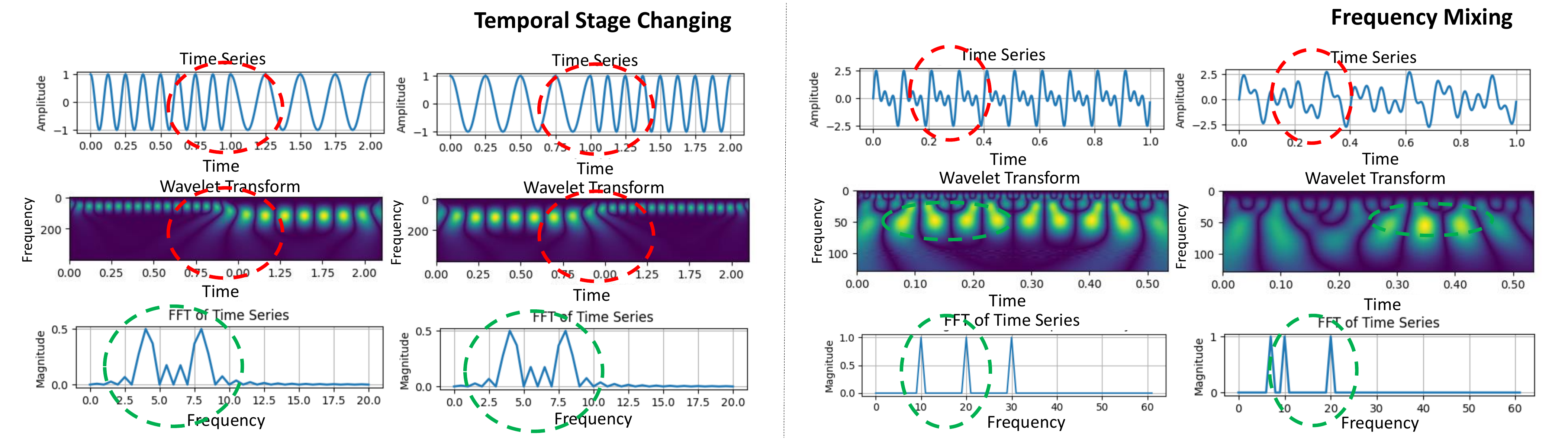}
    \caption{\textbf{Two types of patterns of interest with three multi-modal feature views.} (raw time series, CWT, and FFT of time series) With a green dash circle indicating patterns that are hard to distinguish and, a red dash circle indicating patterns that are easy to identify. The feature encoders take these multi-modal features as input. Difficulties for them to capture certain patterns vary with different transforms, causing different probabilities for them in determining two sample points to be identical.}
    \label{fig:contrast}
\end{figure*}

Eq~\ref{eq:graph_weight} defines a global alignment subgraph $\mathcal{G}$ for low-dimensional reconstruction, where the adjacency function $A_{T_{(\mathcal{X}_i)}, T_{(\mathcal{X}_j)}}$ identifies data points $\mathcal{X}_i, \mathcal{X}_j$ originating from the same sample via transforms $T_{(\mathcal{X}_i)}$ and $T_{(\mathcal{X}_j)}$, which generate modality-specific patterns. Section~\ref{sec:desi} details two edge types and their roles in shaping representation utility, while Fig.~\ref{fig:contrast} visualizes how transforms globally adjust graph weights through local alignment effects.

\textbf{Graph embeddings as entities representations.} For the extracted feature vectors $\mathcal{Z} \in \mathbb{R}^{(k+1)N \times d_z}$, which is stacked by $Z^{(i)}, i\in [0, k]$. We learn them as graph embeddings of the entities by solving Eq~\ref{eq:LE}, which is encoded from the encoder $\phi_\theta^T(X_{\text{\textit{data}}})$ with parameters $\theta$. 

\begin{equation}\label{eq:LE}
\begin{aligned}
   & \min_{\mathcal{Z}: \mathcal{Z}^TD\mathcal{Z}=I} Tr(\mathcal{Z}^T(D- w)\mathcal{Z}) \\
   = & \min_{\mathcal{\theta}: \phi_\theta^T(X_{\text{\textit{data}}})D\phi_\theta(X_{\text{\textit{data}}})=I} Tr(\phi_\theta^T(X_{\text{\textit{data}}})(D- w)\phi_\theta(X_{\text{\textit{data}}})) \\
   & \text{where} \ \ \ \ \  D = \text{diag}(\sum_j w_{ij})
\end{aligned}
\end{equation}

\textbf{Generalization to unseen data.} This can be solved as a generalized eigenvalue problem~\cite{belkin2003laplacian}. To generalize the discussion to unseen data, we further discuss how to generalize the problem to extrapolate to infinite new samples sampled according to the data distribution.

\textbf{Preserving time series encoder only.} Finally, only the shapelet neural encoder~\cite{liang2023contrastive} for the raw MTS data is preserved during adaptations for downstream tasks and inferring processes with the following reasons. (1) \textit{Inference efficiency with a simple structure.} (2) \textit{Universality as MTS as primary input.} This raw data feature encoder memorizes algorithmic knowledge from the transforms and encoders. The scalability of the method is also promoted by the small scale of the neural architecture during inference. 

\section{Regularization Based Multi-Modal Feature Alignment Algorithm}\label{sec:algo}

In this section, we aim to build embeddings that can capture all of the inductive biases introduced, which are learned from a global alignment graph $\mathcal{G}$. 

Initially, we analyze the approximated eigenfunctions of the graph Laplacian operator, each of which arrages the representations according to the transform properties. Subsequently, we introduce training goals to acquiring representations of the input data from these approximated eigenfunctions. We then design an algorithm to minimize the training objective.


\subsection{Training Objective Approximating GLE}\label{sec:desi}

\textbf{Supervisory goal.} The supervisory goal can be seen as learning \textit{graph Laplacian embedding (GLE)}~\cite{trillos2021geometric}, which gives global information about the data manifold. Since we treat the neural encoders as continuous mappings, they automatedly arrange representations according to local fluctuations. As is shown in \textbf{Fig.}~\ref{fig:contrast}, different transforms provide distinct local properties that would saliently influence the local arrangement. Initially, we establish fundamental concepts closely associated with the GLE problem. Subsequently, we introduce the training objectives.

\textbf{Extrapolated distribution.} Training and testing can be seen as independently and uniformly sample data samples with a distribution $p_{\text{\textit{data}}}^\prime: \mathcal{S}^{\text{\textit{data}}} \to \mathbb{R}$ as entities of interest. The training procedure can be seen as sampling jointly from a probability distribution which can approximate and match the equivalence of pairs of the entities, $p_{\text{\textit{sim}}}$, whose possibility density function (PDF) can be constructed as a piecewise function, $ \pi = \frac 1 k p_{T_{(x_i)}, T_{(x_j)}}(x_i, x_j): \mathcal{S}^{\text{\textit{data}}} \times \mathcal{S}^{\text{\textit{data}}} \to [0,1]$, where $T_{(x_m)}=T^{(n)}$ denotes the transform used to acquire $x_m \in X^{(n)}$, and $\mathcal{S}^{\text{\textit{data}}} = \sqcup_{i \in [0,k]} \text{Im}(T^{(i)})$ which we fist introduce in \textbf{Sec.}~\ref{sec:neur_encoder_main}. Since the different transform pairs are sampled equally, the joint distributions for them are simply sliced and normalized. 

\textbf{Capturing distributions with eigenfunctions.} Next, we can generalize $\mathcal{G}$ into a graph describing any of the entities conforms $p_{\text{\textit{data}}}^\prime$, and weighted edges between them. Since the GLE problem is strongly related to the concept of the Laplacian operator and its eigenfunction, we next define them in Def.~\ref{def:laplace} and Def.~\ref{def:lap_eig}, respectively.


\begin{definition}\label{def:laplace}
For a function $f: \mathcal{S}^{\text{\textit{data}}} \to \mathbb{R}$, the \emph{Laplacian operator} $\mathbb{L}$ for PDF $\pi$ is defined as:

\begin{equation}\label{eq:laplacian}
    \begin{aligned}
\mathbb{L}(f)(x)  = f(x) + \int_{x^\prime} \frac {\frac 1 k p_{T_{(x)}, T_{(x^\prime)}}(x, x^\prime)} {p_{\text{\textit{data}}}^\prime(x)} f(x^\prime) dx^\prime \\
  \  = f(x) + \frac 1 k \sum_{i=1}^k \int_{x^\prime \in T^{(i)}(\overline{\mathcal{X}})} \frac {p_{T_{(x)}, T_i}(x, x^\prime)} {p_{\text{\textit{data}}}^\prime(x)} f(x^\prime) dx^\prime    
    \end{aligned}
\end{equation}
\end{definition}

\begin{definition}\label{def:lap_eig}
An \emph{eigenfunction} of the Laplacian operator $\mathbb{L}$ with eigenvalue $\lambda$ is a function $f$ satisfies Eq.~\ref{eq:eigen}:

\begin{equation}\label{eq:eigen}
    \mathbb{E}_{x\sim p_{\text{\textit{data}}}^\prime}[(\mathbb{L}(f)(x) - \lambda f(x))^2]=0 
\end{equation}
\end{definition}

Generally, eigenfunctions with small eigenvalues would correspond to clusters that are almost disconnected from the rest of the graph, which is well-known in the spectral graph theory~\cite{trevisan2017lecture}. 

Based on this and the Laplacian operator $\mathbb{L}$ definition, \textbf{MMFA} enhances base encoder representations. The chosen transform-encoder compositions construct sub-graphs from raw MTS, each revealing distinct discriminative patterns. The enhancement occurs through two mechanisms. The transform-encoder compositions show the robustness of irrelevant patterns that are introduced by each other. The alignment method we proposed assigns high priority to learning these transform properties. 

\textbf{Reinforcing various subtle invariance and patterns.} Some of the important patterns may be ignored by certain compositions but can be enhanced by another one. The added edges help to inject deduced invariance from one composition to the others. These added edges are introduced by $\pi$. \textbf{Theorem}~\ref{the:seman_sim} models more connected components of the graph to eigenvectors by reducing the distance between representations of entities at each end of the edges.

\begin{theorem}\label{the:seman_sim}
Equivalence between Eigenvalues and Distance Reduction of Spectral Embeddings.

$\mathbb{E}_{(x, x^\prime)\sim p_{\text{\textit{sim}}}} [(f(x) - f(x^\prime))^2]$ denotes the expected squared difference between representations of data point pairs under distribution $p_{\text{\textit{sim}}}$, and $f$ is an eigenfunction of $\mathbb{L}$. Eq.~\ref{eq:distance} demonstrates the relationship as follows:

\begin{equation}\label{eq:distance}
    \mathbb{E}_{(x, x^\prime)\sim p_{\text{\textit{sim}}}} [(f(x) - f(x^\prime))^2] = 2\lambda\mathbb{E}_{x \sim p_{\text{\textit{data}}}^\prime}[f(x)^2]
\end{equation}

\end{theorem}

Proof of the theorem is in \textbf{Sec.}~\ref{proof:eq}.

\begin{remark}
\textbf{Theorem}~\ref{the:seman_sim} illustrates that a lower eigenvalue $\lambda$ correlates with reduced distance between nearly aligned pairs. Consequently, minimizing the distance during training aligns the encoder with the eigenfunctions of the Laplacian operator possessing small eigenvalues, effectively segregating the graph-captured relations into relatively distinct clusters.
\end{remark}

We denote raw MTS samples and their distribution as $x \sim p_{\text{\textit{data}}}$. $X$ is a set of samples sampled with $p_{\text{\textit{data}}}$. To capture the global invariance implied in $w$, the corresponding loss can be estimated and bounded by Inequation~\ref{eq:invprime}, where $Z \in \mathbb{R}^{(k+1)\times N \times d }$ which is reshaped from $\mathcal{Z}$, and $z_i^{(m)} \in Z^{(m)}$ denotes the representation for the $i$th sample transformed and encoded by $T^{(m)}$ and $f^{(m)}$. 

To ensure the feasibility of training, we provide an upper bound of the originally proposed loss function to reduce the training workload, which is demonstrated at \textbf{Theorem}~\ref{the:inv_est}.

\begin{theorem}\label{the:inv_est}
Multi-modal Invariance Estimation.

$x \sim p_{\text{\textit{data}}}$ denotes raw Multivariate Time Series (MTS) samples, where $X$ represents a group of samples sampled with $p_{\text{\textit{data}}}$. Strictly, $p_{T^{(i)}, T^{(j)}}(T^{(i)}(x), T^{(j)}(x)) = 1, x \in X$. This captures the global equivariance implied in $w$, the corresponding loss, denoted by $\mathcal{L}_{\text{inv}}^\prime(Z)$, can be estimated and bounded as shown in Inequality $\ref{eq:invprime}$.

\begin{equation}\label{eq:invprime}
\begin{aligned}
    \mathcal{L}_{\text{inv}}^\prime(Z) & = \frac{1}{Nk(k+1)}\sum_{\substack{m=0 \\ n=0 \\ m \neq n}}^{ k} \sum_{i=1}^N ||z_i^{(m)} - z_i^{(n)}||^2_2 \\
    & \leq \frac{2}{N(k+1)} \sum_{i=1}^N\sum_{m=1}^k ||z^{(0)}_i - z^{(m)}_i||^2_2
\end{aligned}
\end{equation}

Proof of the theorem is in \textbf{Sec.}~\ref{proof:inv}.
\end{theorem}
The expression \( \mathcal{L}_{\text{inv}}^\prime(Z) \) simplifies the optimization objective aimed at capturing global equivariance among the multi-modal features.

According to the \textbf{Theorem}~\ref{the:inv_est}, we define $\mathcal{L}_{inv}(Z)$ as the optimization objective capturing global invariance among the multi-modal features.

\begin{equation}\label{eq:inv}
    \mathcal{L}_{inv}(Z) = \frac{1}{N(k+1)} \sum_{i=1}^N \sum_{m=1}^k ||z^{(0)}_i - z^{(m)}_i||^2_2
\end{equation}

\begin{theorem}\label{the:orth}
Orthogonality of Eigenfunction-Based Representations.

For $ \forall g, h: \mathbb{R^{\text{\textit{data}}}} \to \mathbb{R}$, $\mathbb{E}{x \sim p_{\text{\textit{data}}}'}[\mathbb{L}(g)(x) \cdot h(x)] =\mathbb{E}{x \sim p_{\text{\textit{data}}}'}[g(x) \cdot \mathbb{L}(h)(x)]$. Therefore, $\mathbb{L}$ is a symmetric linear operator. The representations are outputs derived from approximated eigenfunctions of $\mathbb{L}$ possessing low eigenvalues, $f_{\text{rep}}(\cdot) = [f_1(\cdot), f_2(\cdot), \dots, f_{d_z}(\cdot)]^T$, and $\mathbb{L}$ is a symmetric real value linear operator. According to spectral theorem~\cite{trevisan2017lecture}, it is feasible to ensure orthogonality while maximizing the overall information:

\begin{algorithm}[t]
   \caption{Asymmetric Neural Encoders Alignment}
   \label{algo}
   \begin{algorithmic}
   \STATE {\bfseries Input:} $D$: MTS dataset; $T^{(i)}, i \in [k]$: transform operators; $f^{(i)}, i \in [k]$: neural encoder architectures.
   \STATE {\bfseries Output:} $f$: Trained ST encoder for MTS representation.
   
   \STATE $\theta_{f} \gets$ Initialize parameters of the raw MTS encoder.
   \FOR{$i \in [k]$}
       \STATE $\theta_{f^{(i)}} \gets$ Initialize parameters of the encoders w.r.t $T^{(i)}$.
   \ENDFOR
   \WHILE{not converged}
       \STATE $X \gets$ Sample a mini-batch from $D$.
       \FOR{$i \in [k]$}
           \STATE $X^{(i)} \gets T^{(i)}(X)$ \# Transform raw MTS.
           \STATE $\theta_{\tilde f}, \theta_{\tilde f^{(i)}} \gets U(f, f^{(i)}, X, X^{(i)})$ \# Accumulate gradients.
           \STATE $\theta_{f^{(i)}} \gets \theta_{f^{(i)}} + \epsilon \frac{1}{ k} (\theta_{\tilde f^{(i)}} - \theta_{f^{(i)}})$ \# Update encoder.
       \ENDFOR
       \STATE $\theta_{f} \gets \theta_{f} + \epsilon (\theta_{\tilde f} - \theta_{f})$ \# Update encoder.
   \ENDWHILE
   \STATE {\bfseries Return:} $f$
   \end{algorithmic}
\end{algorithm}

\begin{equation}
\mathbb{E}_{x \sim p_{\text{\textit{data}}}'}[f_{\text{rep}}(x)f_{\text{rep}}(x)^T] = \mathbb{I}
\end{equation}
\end{theorem}

According to \textbf{Theorem}~\ref{the:orth}, we design two relevant losses, $\mathcal{L}_{var}$ and $\mathcal{L}_{cov}$, in Eq.~\ref{eq:var} and Eq.~\ref{eq:cov}, respectively.

\begin{equation}\label{eq:var}
\begin{aligned}
    & \mathcal{L}_{cov}(Z) \\
    & = \frac 1 {(N - 1)kd} \sum_{l=1}^k \sum_{m \neq n} [\sum_{i=1}^N(z_{i}^{(l)} - \bar z^{(l)})(z_{i}^{(l)} - \bar z^{(l)})^T]_{m,n} \\ 
    & where \ \bar z^{(l)} = \frac 1 n \sum_{i=1}^n z_{i}^{(l)}
\end{aligned}
\end{equation}

\begin{equation}\label{eq:cov}
    \mathcal{L}_{var}(Z) = \frac 1 {dk} \sum_{l=0}^k \sum_{j=1}^d max(0, 1 - \sqrt{Var(\{z^{(l)}_{i,j} | i\in [N]\}) + \epsilon})
\end{equation}
According to Eq.~\ref{eq:inv}, Eq.~\ref{eq:var} and Eq.~\ref{eq:cov}, the end-to-end training object is obtained in equation~\ref{eq:loss}. $\alpha$, $\beta$ and $\gamma$ are empirical parameters.

\begin{equation}\label{eq:loss}
    \mathcal{L} = \alpha \mathcal{L}_{cov} + \beta \mathcal{L}_{var} + \gamma \mathcal{L}_{inv}
\end{equation}

\textbf{Spectral selection mechanism ensures the balance of transforms as information sources.} Our further discussion deduced the closed-form solutions of linear degenerated \textbf{MMFA} with $k=0$ in \textbf{Lemma}~\ref{lem:PCA}, $k=1$ in \textbf{Lemma}~\ref{lem:CCA}, and $k=p > 2 $ in \textbf{Lemma}~\ref{lem:multiCCA} transforms aligned with the identity transformation in \textbf{Sec.}~\ref{sec:recover}.  According to the closed-form optimal solutions, we can observe that \textbf{MMFA} serves as a spectral selection mechanism for balancing variances of empirical distributions measured with the transformed features.

\begin{theorem}\label{theorem:recover}
    The optimization problem of \textbf{MMFA} recovers kernel principal component analysis (KPCA)~\cite{kpca} and kernel canonical correlation analysis (KCCA)~\cite{fukumizu2007statistical}.
\end{theorem}

Proof of the theorem is in \textbf{Sec.}~\ref{sec:recoverk}.

\begin{table*}[t]
    
    \caption{Comparisons of performance in MTS classification, evaluated on accuracy scores. The top-performing results of URL methods are shown in bold, with $\dag$ denoting the overall best performance.}
    \small
    \label{tab:classification}
    \resizebox{1\textwidth}{!}{
    \begin{tabular}{lcccccc|cccccccc}
    \toprule
    \multirow{2}{*}{\textbf{Dataset}} & \multicolumn{6}{c|}{\textbf{Tailored Classification Approaches}} & \multicolumn{8}{c}{\textbf{Unsupervised Representation Learning + Classifier}} \\
    \cline{2-15}

    & \method{DTWD}{bagnall2018uea} 
& \method{MLSTM-FCNs}{karim2019multivariate} 
& \method{TapNet}{zhang2020tapnet} 
& \method{ShapeNet}{li2021shapenet} 
& \method{OSCNN}{tang2020omni} 
& \method{DSN}{xiao2022dynamic} 
& \method{TS2Vec}{yue2022ts2vec} 
& \method{T-Loss}{franceschi2019unsupervised} 
& \method{TNC}{tonekaboni2021unsupervised} 
& \method{TS-TCC}{eldele2021time} 
& \method{TST}{zerveas2021transformer} 
& \method{CSL}{liang2023contrastive}
 &  MMFA-aug & \textbf{MMFA}  \\
    \midrule
    	
ArticularyWordRecognition   & 0.987   & 0.973        & 0.987    & 0.987         & 0.988        & 0.984         & 0.987                 & 0.943          & 0.973        & 0.953            & 0.977  & 0.990    &  0.973 & \textbf{0.993}$^\dag$\\
AtrialFibrillation   & 0.200     & 0.267        & 0.333    & 0.400	        & 0.233	       & 0.067         & 0.200	               & 0.133	        & 0.133	       & 0.267	          & 0.067	  &  0.533    & 0.533 &  \textbf{0.600}$^\dag$\\
BasicMotions   & 0.975	  & 0.950	    & 1.000$^\dag$ & 1.000$^\dag$      & 1.000$^\dag$     & 1.000$^\dag$      & 0.975	             &\textbf{1.000}$^\dag$	& 0.975	       & \textbf{1.000}$^\dag$& 0.975	& \textbf{1.000}$^\dag$ & \textbf{1.000}$^\dag$      & \textbf{1.000}$^\dag$ \\
CharacterTrajectories  & 0.989	  & 0.985	     & 0.997	& 0.980	        & 0.998$^\dag$ & 0.994         & \textbf{0.995}        & 0.993	        & 0.967	       & 0.985	          & 0.975	 & 0.991    & 0.987     & 0.991 \\
Cricket   &1.000$^\dag$ & 0.917      	 & 0.958	& 0.986	        & 0.993	       & 0.989         & 0.972	               & 0.972	        & 0.958	       & 0.917	      &\textbf{1.000}$^\dag$  & 0.994  & 0.986  & \textbf{1.000}$^\dag$  \\
DuckDuckGeese   & 0.600     & 0.675	     & 0.575	& 0.725$^\dag$  & 0.540	       & 0.568         & \textbf{0.680}         & 0.650	        & 0.460	       & 0.380	          & 0.620	& 0.380    & 0.620   & \textbf{0.680}            \\
EigenWorms  & 0.618	  & 0.504	   	 & 0.489	& 0.878$^\dag$	& 0.414        & 0.391         & \textbf{0.847}        & 0.840	        & 0.840	       & 0.779	          & 0.748	& 0.779   & 0.768    & 0.840           \\
Epilepsy   & 0.964	  & 0.761	    	 & 0.971	& 0.987	        & 0.980	       & 0.999$^\dag$  & 0.964	        & 0.971	        & 0.957	       & 0.957	          & 0.949  & 0.986	 & 0.899     & \textbf{0.987}$^\dag$ \\
ERing   & 0.133	  & 0.133	    	 & 0.133	& 0.133	        & 0.882	       & 0.922         & 0.874	               & 0.133	        & 0.852	       & 0.904	          & 0.874	 & \textbf{0.967}$^\dag$   &  0.922  & \textbf{0.967}$^\dag$  \\
EthanolConcentration  & 0.323	  & 0.373	   	 & 0.323	& 0.312	        & 0.241	       & 0.245         & 0.308	               & 0.205	        & 0.297	       & 0.285	          & 0.262	 & 0.498    &       0.331  &  \textbf{0.551}$^\dag$ \\
FaceDetection  & 0.529	  & 0.545	    	 & 0.556	& 0.602	        & 0.575	       & 0.635$^\dag$  & 0.501	               & 0.513	        & 0.536	       & 0.544	          & 0.534	& \textbf{0.593}  & 0.526    & 0.555        \\
FingerMovements  & 0.530	  & 0.580	    	 & 0.530	    & 0.580	        & 0.568	       & 0.492         & 0.480	               & 0.580	        & 0.470	       & 0.460	          & 0.560	 & 0.590  & 0.490    & \textbf{0.610}$^\dag$  \\
HandMovementDirection  & 0.231	  & 0.365	   	 & 0.378	& 0.338	        & 0.443 & 0.373         & 0.338	               & 0.351	        & 0.324	     & 0.243	          & 0.243    &    0.432   &  0.351 &
 \textbf{0.487}$^\dag$\\
Handwriting  & 0.286	  & 0.286	    	 & 0.357	& 0.451	        & 0.668$^\dag$  & 0.337         & 0.515	               & 0.451	        & 0.249	       & 0.498	          & 0.225	& \textbf{0.533}   & 0.370     & 0.472    \\
Heartbeat  & 0.717	  & 0.663	    	 & 0.751	& 0.756	        & 0.489	       & 0.783$^\dag$  & 0.683	               & 0.741	        & 0.746	       & 0.751	  & 0.746	& 0.722      &  0.720 &   \textbf{0.761} \\

InsectWingbeat  & N/A	      & 0.167	   	 & 0.208	& 0.250	        & 0.667$^\dag$ & 0.386         & 0.466                 & 0.156	        &\textbf{0.469} & 0.264	          & 0.105	 & 0.256      & 0.425    & \textbf{0.469}       \\  

JapaneseVowels  & 0.949	  & 0.976	    	 & 0.965	& 0.984	        & 0.991$^\dag$ & 0.987         & 0.984	               & \textbf{0.989} & 0.978	       & 0.930	          & 0.978 & 0.919	 &  0.960  & 0.978     \\
Libras & 0.870	  & 0.856	   	 & 0.850	    & 0.856	        & 0.950	       & 0.964$^\dag$  & 0.867	               & 0.883	        & 0.817	       & 0.822	          & 0.656	& \textbf{0.906} & 0.833    & 0.894   \\
LSST  & 0.551	  & 0.373	    	 & 0.568    & 0.590	        & 0.413	       & 0.603         & 0.537	               & 0.509	        & 0.595	       & 0.474	          & 0.408	& 0.617    & 0.372     & \textbf{0.622}$^\dag$    \\
MotorImagery & 0.500	  & 0.510	    	 & 0.590	    &0.610$^\dag$    & 0.535	       & 0.574         & 0.510	               & 0.580	        & 0.500	       & 0.610  & 0.500	  &  0.610 & 0.510   & \textbf{0.630}$^\dag$  \\
NATOPS  & 0.883	  & 0.889	    	 & 0.939	& 0.883	        & 0.968	       & 0.978$^\dag$        & 0.928        & 0.917          & 0.911	       & 0.822	          & 0.850	& 0.878     & 0.900     & \textbf{0.933}         \\
PEMS-SF  & 0.711	  & 0.699	   	 & 0.751	& 0.751	        & 0.760	       & 0.801         & 0.682	               & 0.676	        & 0.699	       & 0.734	          & 0.740	& 0.827      & 0.827      & \textbf{0.838}$^\dag$ \\

PenDigits  & 0.977	  & 0.978	     & 0.980	    & 0.977	        & 0.986	       & 0.987         & 0.989	               & 0.981	        & 0.979	       & 0.974	          & 0.560 & \textbf{0.990}$^\dag$	& 0.979   & 0.980    \\
PhonemeSpectra  & 0.151	  & 0.110	    	 & 0.175	& 0.298	        & 0.299$^\dag$ & 0.320          & 0.233	               & 0.222	        & 0.207	       & 0.252	          & 0.085	& \textbf{0.255} & 0.183     & 0.216  \\
RacketSports  & 0.803	  & 0.803	    	 & 0.868	& 0.882$^\dag$	& 0.877	       & 0.862         & 0.855	               & 0.855	        & 0.776	       & 0.816	          & 0.809  & \textbf{0.882}$^\dag$  & 0.815   & 0.862 \\
SelfRegulationSCP1  & 0.775	  &0.874		 & 0.652	& 0.782	        & 0.835	       & 0.717         & 0.812	               & 0.843	        & 0.799	       & 0.823	          & 0.754 &  0.846 &  0.887	    &  \textbf{0.901}$^\dag$  \\
SelfRegulationSCP2  & 0.539	  & 0.472	    	 & 0.550	    & 0.578     	& 0.532	       & 0.464         & 0.578            & 0.539	        & 0.550	       & 0.533	          & 0.550 & 0.496	 & 0.617     &  \textbf{0.628}$^\dag$  \\
SpokenArabicDigits  & 0.963	  & 0.990	    	 & 0.983	& 0.975	        & 0.997$^\dag$ & 0.991         & 0.988	               & 0.905	        & 0.934	       & 0.970	          & 0.923	& \textbf{0.990}    &  0.980   & \textbf{0.990} \\
StandWalkJump  & 0.200	  & 0.067      	 & 0.400	    & 0.533	        & 0.383	       & 0.387         & 0.467	               & 0.333	        & 0.400	       & 0.333	          & 0.267	     & \textbf{0.667}$^\dag$  & \textbf{0.667}$^\dag$  &  \textbf{0.667}$^\dag$ \\
UWaveGestureLibrary  & 0.903	  & 0.891	      & 0.894	& 0.906	        & 0.927 & 0.916         & 0.906	               & 0.875	        & 0.759	       & 0.753	          & 0.575	& 0.922    & 0.922  &  \textbf{0.931}$^\dag$ \\
 
\hline
\makecell[l]{\textbf{Avg Cohen's d Effect}}  \\
incl. InsectWingbeat & /	  & -0.670	    	 & -0.024	& 0.366	        & 0.357	       & 0.280         & 0.089  & -0.176 & -0.450 & -0.447 & -0.860  & 0.502 & 0.096 & 
\textbf{0.937}$^\dag$  \\

excl. InsectWingbeat & -0.376	  & -0.631	   & 0.033	& 0.434	        & 0.321	       & 0.308         & 0.094  & -0.116 &  -0.473 & -0.425 & -0.825  & 0.567 & 0.109 & \textbf{0.979}$^\dag$  \\
\hline
\makecell[l]{\textbf{Avg Ranking} (URL only)} & / & / & / & / & / & / & 3.96 & 4.63 & 5.40  & 5.30 &  6.20 & 2.83 & 4.43 & \textbf{1.60} \\
\makecell[l]{\textbf{Avg Ranking} (All)} & 
9.40 & 9.43 & 6.90 &  5.43 & 5.37 &
   5.83 & 6.83 & 7.90 & 9.37  & 9.10 &
       10.46 & 4.40 & 7.17  & \textbf{2.63}$^\dag$ \\
    \bottomrule
    \end{tabular}
    }
\end{table*}

\subsection{Asymmetric Encoders Alignment Optimizing Algorithm}

We finally design Algorithm~\ref{algo} minimizing the training objective by mini-batch gradient descent. Algorithm~\ref{algo} takes the MTS dataset $D$, a set of transform operators $T^{(i)}$, and corresponding neural architecture for the encoders as input. Then the parameters of the encoders $f$ and $f^{(i)}$ are initialized in lines 1-3. In lines 7-9, update step $U$ is defined as $U(f, f^{(i)}, X, X^{(i)}) = \theta - \nabla_\theta \mathcal{L}([f^{(i)}(X^{(i)}), f(X)])$, where $\theta$ denotes the parameters of the two encoders. We divide the designed training objective into $k$ parts, iteratively computes and accumulates gradient for both $f$ and $f^{(i)}$. After accumulating all the gradients of the batch, the encoder $f$ is updated.


After the training phase, we only need to preserve the encoder, $f$, for raw MTS and adapt it to diverse downstream tasks.

\section{Empirical Analysis}\label{sec:empi}

In this section we provide empirical evidence to justify our main claims. To ensure throughness and fairness of the analysis, we provide details about the implementations of each benchmarks and baselines in \textbf{Sec.}~\ref{sec:exp_detail}

\textbf{Extensive experiments on various datasets.} In this section, we conduct extensive experiments containing 31 real-world datasets for three important downstream tasks, classification, clustering, and anomaly detection. The 30 datasets for classification and 12 of them for clustering vary with their \textbf{training set sizes}, \textbf{dimensionalities}, \textbf{lengths}, and \textbf{numbers of classes} which are detailed in \textbf{Tab.}~\ref{tab:uea_statistics}. The remaining 1 datasets for anomaly detection have different \textbf{numbers of entities}, \textbf{dimensionality}, \textbf{length}, and \textbf{anomaly ratio} which are detailed in \textbf{Tab.}~\ref{tab:ad_statistics}. Standard metrics are used for various tasks, i.e., Accuracy (Acc)~\cite{bagnall2018uea} for classification, Cohen's d~\cite{becker2000effect} for standardizing performance on various datasets, RI~\cite{Hubert1985ComparingP} and NMI~\cite{Strehl2002ClusterE} for clustering, and F1-score~\cite{li2021multivariate} for anomaly detection.

\textbf{Baselines with significant diversity.} The baselines are divided into two groups,i.e., the URL methods and methods tailored to specific tasks. All the URL methods share the same evaluation protocol on downstream tasks, by feeding the representations to classical models, i.e., SVM~\cite{Cortes1995SupportVectorN}, IF~\cite{Liu2008IsolationF}, and KMeans~\cite{MacQueen1967SomeMF}. The tailored methods are evaluated following the original implementations. More detailed demonstrations of the downstream tasks, datasets, evaluation metrics, and baselines are elaborated in \textbf{Sec.}~\ref{sec:exp_detail}.


\textbf{Comprehensive research questions.} We evaluate the robustness and adaptability of \textbf{MMFA} across diverse datasets and domains. Through experiments, we aim to scrutinize algorithms, assess scalability, and compare them to the proposed \textbf{MMFA} framework, outlined by our research questions (RQs).

\begin{table*}[h]
    \centering
    \caption{The MTS clustering performance evaluated on RI~\cite{Hubert1985ComparingP} and NMI~\cite{Strehl2002ClusterE}, with the most effective results from URL methods in bold, with $\dag$ indicating superiority over others.}
    \label{tab:clustering}
    \resizebox{1\linewidth}{!}{
    \begin{tabular}{lcccccccc|cccccccc}
    \toprule
    \multirow{2}{*}{\textbf{Dataset}} & \multirow{2}{*}{\textbf{Metric}} & \multicolumn{7}{c|}{\textbf{Tailored Clustering Approaches}} & \multicolumn{6}{c}{\textbf{Unsupervised Representation Learning + Clustering}} \\
    \cline{3-17}&
      & \method{MC2PCA}{li2019multivariate} & \method{TCK}{mikalsen2018time} & \method{m-kAVG+ED}{ozer2020discovering} & \method{m-kDBA}{ozer2020discovering} & \method{DeTSEC}{ienco2020deep} & \method{MUSLA}{zhang2022multiview} & \method{Time2Feat}{bonifati2022time2feat} & \method{TS2Vec}{yue2022ts2vec} & \method{T-Loss}{franceschi2019unsupervised} & \method{TNC}{tonekaboni2021unsupervised} & \method{TS-TCC}{eldele2021time} & \method{TST}{zerveas2021transformer} & \method{CSL}{liang2023contrastive} & MMFA-aug & \textbf{MMFA} \\
    \midrule
\multirow{2}{*}{ArticularyWordRecognition} & RI	& 0.989 & 0.973 &	0.952 &	0.934 &	0.972  &	0.977 &   0.977 &	0.980 &	0.975 &	0.938 &	0.946 &	0.978 & 0.990 &	0.990  & \textbf{0.998}$^\dag$     \\
                                           & NMI & 0.934 &	0.873 &	0.834 &	0.741 &	0.792  &	0.838 & 0.881 &	0.880 &	0.842 &	0.565 &	0.621 &	0.866 &	0.942 & 0.942& \textbf{0.983}$^\dag$       \\
\multirow{2}{*}{AtrialFibrillation} & RI	& 0.514 & 0.552 &  	0.705 &	0.686 &	0.629  &	0.724 & 0.600 &  0.465 &	0.469 &	0.518 &	0.469 &	0.444 &	\textbf{0.743}$^\dag$  &	0.524  & 0.648  \\
                                           & NMI & 0.514	& 0.191 &  	0.516 &	0.317 &	0.293  &	0.538 &	0.261 & 0.080 &	0.149 &	0.147 &	0.164 &	0.171 &	\textbf{0.587}$^\dag$& 0.251 & 0.305    \\
\multirow{2}{*}{BasicMotions} & RI & 0.791	& 0.868 &  	0.772 &	0.749 &	0.717  &	1.000$^\dag$ & 0.883 &	0.854 &	0.936 &	0.719 &	0.856 &	0.844 & \textbf{1.000} &	0.845    &  \textbf{1.000}      \\
                                           & NMI & 0.674	& 0.776 &  	0.543 &	0.639 &	0.800  &	1.000$^\dag$ & 0.833 &	0.820 &	0.871 &	0.394 &	0.823 &	0.810 & \textbf{1.000} & 0.806 & \textbf{1.000}    \\
\multirow{2}{*}{Epilepsy} & RI & 0.613	& 0.786 &  	0.768 &	0.777 &	0.840  & 0.816 &	0.897$^\dag$ & 0.706 &	0.705 &	0.650 &	0.736 &	0.718 & 0.873  &	0.679    & \textbf{0.876}   \\
                                           & NMI & 0.173	& 0.534 &  	0.409 &	0.471 &	0.346  &	0.601 &	0.814$^\dag$ & 0.312 &	0.306 &	0.156 &	0.451 &	0.357 & 0.705  & 0.530  & \textbf{0.721}    \\
\multirow{2}{*}{ERing} & RI	& 0.756 & 0.772 &  	0.805 &	0.775 &	0.770  &	0.887  & 0.841  & 0.925 &	0.885 &	0.764 &	0.821 &	0.867 & 0.968  &	0.968  & \textbf{0.972}$^\dag$  \\
                                           & NMI & 0.336	& 0.399 &  	0.400 &	0.406 &	0.392  &	0.722 &	0.641 &0.775 &	0.672 &	0.346 &	0.478 &	0.594 &	0.906 & 0.906 & \textbf{0.918}$^\dag$   \\
\multirow{2}{*}{HandMovementDirection} & RI & 0.627	& 0.635 &  	0.697 & 0.685 &	0.628  &	0.719$^\dag$ & 0.577 & 0.609 	& 0.599 & 0.613 &	0.608 &	0.607 & 0.651 &	0.630 & \textbf{0.657} \\
                                           & NMI & 0.067	& 0.103 &  	0.168 &	0.265 &	0.112  &	0.398$^\dag$ & 0.062 &	0.044 &	0.034 &	0.051 &	0.053 &	0.039 &	0.175 & 0.118  &	\textbf{0.185} \\
\multirow{2}{*}{Libras} & RI & 0.892 	& 0.917 &  	0.911 &	0.913 &	0.907  &	0.941 &	0.926 & 0.904 &	0.922 &	0.896 &	0.881 &	0.886 & 0.941 &	0.938 & \textbf{0.944}$^\dag$ \\
                                           & NMI	& 0.577 & 0.620 &  	0.622 &	0.622 &	0.602  &	0.724 &	0.703 & 0.542 & 0.654 &	0.464 &	0.373 &	0.400 &	0.761 & 0.730 & \textbf{0.785}$^\dag$    \\
\multirow{2}{*}{NATOPS} & RI & 0.882	& 0.833 &  	0.853 &	0.876 &	0.714  &	0.976$^\dag$ & 0.777 &	0.817 &	0.836 &	0.700 &	0.792 &	0.809 &	0.876 &	0.850  &  \textbf{0.919}    \\
                                           & NMI & 0.698	& 0.679 &  	0.643 &	0.643 &	0.043  &	0.855$^\dag$ & 0.482 &	0.523 &	0.558 &	0.222 &	0.513 &	0.565 &	0.657 &	0.628   & \textbf{0.864} \\
\multirow{2}{*}{PEMS-SF} & RI & 0.424	& 0.191 &  	0.817 &	0.755 &	0.806  &	0.892$^\dag$ & 0.818 &	0.765 &	0.746 &	0.763 &	0.789 &	0.726 &	0.858 & 0.842 & \textbf{0.876}     \\
                                           & NMI & 0.011	& 0.066 &  	0.491 &	0.402 &	0.425  &	0.614$^\dag$ & 0.495	& 0.290 &	0.102 & 0.278 &	0.331 &	0.026 &	0.537 & 0.528 &  \textbf{0.583} \\
\multirow{2}{*}{PenDigits} & RI	& 0.929 & 0.922 &  	0.935 &	0.881 &	0.885  &	0.946 & 0.914 &	0.941 &	0.936 &	0.873 &	0.857 &	0.818 & \textbf{0.950} &	0.921  & 0.932       \\
                                           & NMI & 0.713	& 0.693 &  	0.738 &	0.605 &	0.563  &	0.826$^\dag$ & 0.705 &	0.776 & 0.749 &	0.537 &	0.339 &	0.090 & \textbf{0.822} & 0.703  & 0.730  \\
\multirow{2}{*}{StandWalkJump} & RI & 0.591	& 0.762 &  	0.733 &	0.695 &	0.733  &	0.771$^\dag$ & 0.500 &	0.410 &	0.410 &	0.457 &	0.589 &	0.467 & 0.724  &	0.591  & \textbf{0.761}    \\
                                           & NMI & 0.350	& 0.536 &  	0.559 &	0.466 &	0.556  &	0.609$^\dag$ & 0.077 &	0.213 &	0.213 &	0.193 &	0.187 &	0.248 &	0.554 & 0.336 & \textbf{0.611}   \\
\multirow{2}{*}{UWaveGestureLibrary} & RI & 0.883	& 0.913 &  	0.920 &	0.893 &	0.879  &	0.913 & 0.863 &	0.865 &	0.893 &	0.817 &	0.796 &	0.779 &	0.927 &	0.927    &  \textbf{0.929}$^\dag$  \\
                                           & NMI & 0.570	& 0.710 &  	0.713 &	0.582 &	0.558  &	0.728 &	0.610 & 0.511 &	0.614 & 0.322 &	0.215 &	0.244 &	0.731 &	0.731 & \textbf{0.749}$^\dag$   \\
\hline     
\multirow{2}{*}{\textbf{Avg Cohen's d Effect}}            & RI & -0.451 	& -0.129 & 	0.287 &  -0.065	& -0.230   &1.096 &-0.012	 &-0.227	 &-0.111	 &-1.095  &-0.771	 &-0.816  & 1.084 &  0.321 & \textbf{1.119}$^\dag$ \\
& NMI	& -0.244 & 0.000 & 	0.217	& 0.004	   & -0.316		& 1.152$^\dag$	& 0.153	& -0.265	& -0.232	& -1.302	&-0.910	& -0.892 & 1.071 & 0.478 & \textbf{1.089}   \\
                                           
\hline
\textbf{Avg Ranking } (URL only) & & / & / & / & / & / &
        / & / &  5.08 & 5.25 &  6.92 & 6.08    &  5.92
       & 1.75 & 3.25 & \textbf{1.33}$^\dag$    \\
\textbf{Avg Ranking } (All) & & 9.92 & 7.83 & 6.75 & 8.50 & 9.42 & 2.92 & 8.17 & 9.25 & 8.92 & 12.67 & 11.25 & 11.67  & 2.66 & 6.33 & \textbf{2.42}$^\dag$ \\
    \bottomrule
    \end{tabular}}
\end{table*}

\begin{table}[t]
        \centering
        \caption{Comparison of performance of on UCR anomaly detection dataset. The sliding window's length is set to 256, with the most optimal outcomes highlighted in bold. }
        \label{tab:ad_ucr}
        \resizebox{\linewidth}{!}{
        \begin{tabular}{lccccccc}
        \toprule
         \textbf{Methods on UCR} & IF-s & \method{TST}{zerveas2021transformer} 
& \method{TS-TCC}{eldele2021time} 
& \method{T-Loss}{franceschi2019unsupervised} 
& \method{TS2Vec}{yue2022ts2vec} 
& \method{CSL}{liang2023contrastive} &\textbf{MMFA}  \\
         \midrule

Precision 
&0.9778
&0.7598
&0.7863
&0.7986
&0.8147
&0.8040
&0.9218 \\

Recall 
&0.0069
&0.3481
&0.3431
&0.3556
&0.3551
&0.3664
&0.4717 \\

F1  
&0.0137
&0.4775
&0.4777
&0.4921
&0.4946
&0.5034
&\textbf{0.6241}\\

Accuracy  
&0.0069
&0.9881
&0.9877
&0.9882
&0.9881
&0.9886
&\textbf{0.9920} \\

          \bottomrule
        \end{tabular}
        }
\end{table}

\begin{researchq}
  How does the \textbf{MMFA} framework perform compared to other unsupervised methods?
  \end{researchq}
  
  \textbf{Tab.}~\ref{tab:classification}, \ref{tab:clustering}, and \ref{tab:ad_ucr} compare classification, clustering, and anomaly detection results. The proposed \textbf{MMFA} consistently outperforms unsupervised baselines across tasks, achieving top accuracy in 21/30 datasets. A variant using only basic data augmentations (MMFA-aug) underperforms significantly, as \textbf{MMFA}'s integration of diverse transforms and neural encoders extracts denoised, discriminative patterns (Sec.~\ref{sec:algo}).  

In clustering, \textbf{MMFA} excels except on PenDigits (too short sequences) and AtrialFibrillation. For anomaly detection, it outperforms competitors by large margins, producing compact representation boundaries through multi-modal feature alignment.  

By leveraging algorithmic biases from multi-modal transforms rather than simple augmentations, \textbf{MMFA} induces robust global structures, surpassing methods reliant on limited feature engineering.
  
  \begin{researchq}
  How does the \textbf{MMFA} framework perform compared to the supervised approaches and approaches tailored for the downstream tasks?
  \end{researchq}
  
  The comparison between \textbf{MMFA} and tailored discriminative models on classification tasks is shown in \textbf{Tab.}~\ref{tab:classification}. \textbf{MMFA} outperforms tailored fully supervised competitors for time series classification on 16 UCR classification datasets, indicating its extraction of richer information directly from raw time series during inference.
  
  According to \textbf{Tab.}~\ref{tab:clustering}, over all of the clustering tasks, among the tailored methods, only MUSLA~\cite{zhang2022multiview} is compatible with \textbf{MMFA}. \textbf{MMFA} also shows significant advantages among the URL methods.
  
  In contrast to the compared tailored methods in \textbf{Tab.}~\ref{tab:classification} and \textbf{Tab.}~\ref{tab:clustering}, unsupervised methods learn to transform properties of data without task-specific tuning, enhancing feature extraction capabilities that go beyond specific tasks, and fostering potential generalizability.

\begin{researchq}
What is the impact of data characteristics?
\end{researchq}

\begin{figure}
  \centering
  \includegraphics[width=0.95\linewidth]{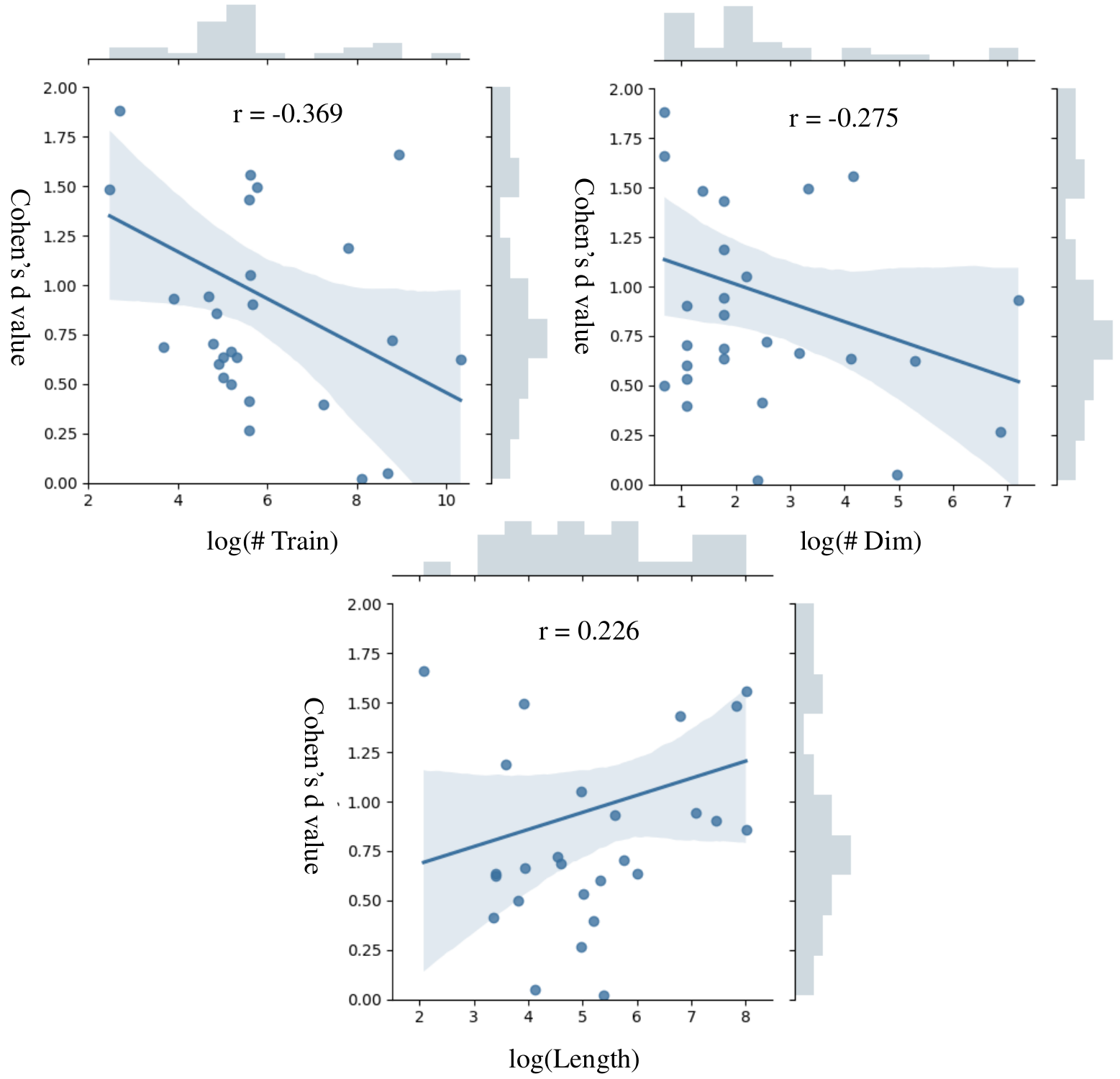}
  \caption{\textbf{Correlation plot} illustrating the relationships between Cohen's d effect sizes of performance improvement made by \textbf{MMFA} on 30 UEA datasets and characteristics of the datasets, i.e., training size, time series dimensionality, and time length. Logarithmic transforms are employed to enhance the linearity of the data.}
  \label{fig:data_char}
\end{figure}

In our analysis, the evaluated datasets exhibit significant variations in training dataset sizes, numbers of channels, time series length, and salient patterns. \textbf{Tab.}~\ref{tab:uea_statistics} illustrates the diversity: training set sizes range from 12 (StandWalkJump, comprising 3 classes with only 4 samples per class, constituting a 3-way, 4-shot learning task) to 30,000 (InsectWingbeat, characterized by abundant training samples). Consequently, we conduct correlation analysis on the UEA datasets focusing on these three data characteristics and show the findings in \textbf{Fig.}~\ref{fig:data_char}.

As depicted in (a) of \textbf{Fig.}~\ref{fig:data_char}, \textbf{MMFA} exhibits a high effect size on performance improvement with smaller training sets. This observation suggests that \textbf{MMFA} excels in achieving higher few-shot performance.

A slight negative correlation exists between classification performance improvement effect size and dataset dimensionality, which shows \textbf{MMFA} outperforms more significantly on low dimensional datasets, showcasing \textbf{MMFA}'s strong generalization effects that avoid overfitting with certain patterns with limited input features.

Conversely, positive correlations between effect size and time series length indicate \textbf{MMFA}'s stronger adaptability to longer time series than baselines. However, a limitation is that \textbf{MMFA} struggles with shorter time series (e.g., PenDigits datasets with $\text{length} = 8$), evident in lower performance across classification and clustering tasks. This limitation may stem from the transforms nature, requiring optimal performance from a time series of specific lengths.

\begin{researchq}
  Ablation (Ab) and leave one out (LOO) study.
\end{researchq}
According to Table~\ref{ablation}, more transformations lead to better performance and lower variation in the magnitude of the overall effect. This is the evidence for the robustness of the model.

\begin{table*}[h]
  \centering
  \caption{Ablation (Ab) and leave one out (LOO) study.}\label{ablation}
  \resizebox{0.8\linewidth}{!}{\begin{tabular}{lcccccccccc}
  \toprule
  Dataset & CGau2 & RP & GADF & DFT & WEASEL & CGau2-loo & RP-loo & GADF-loo & DFT-loo & WEASEL-loo \\
  \midrule
  ArticularyWordRecognition & 0.983 & 0.983 & 0.980 & 0.980 & 0.980 & 0.983 & 0.980 & 0.983 & 0.987 & 0.983 \\
  AtrialFibrillation & 0.533 & 0.533 & 0.533 & 0.467 & 0.467 & 0.533 & 0.533 & 0.533 & 0.533 & 0.533 \\
  BasicMotions & 1.000 & 1.000 & 1.000 & 1.000 & 1.000 & 1.000 & 1.000 & 1.000 & 1.000 & 1.000 \\
  CharacterTrajectories & 0.991 & 0.987 & 0.987 & 0.991 & 0.989 & 0.990 & 0.991 & 0.988 & 0.990 & 0.991 \\
  Cricket & 0.958 & 0.944 & 0.958 & 0.944 & 0.958 & 0.972 & 0.972 & 0.944 & 0.958 & 0.972  \\
  DuckDuckGeese & 0.300 & 0.360 & 0.340 & 0.340 & 0.360 & 0.320 & 0.300 & 0.280 & 0.340 & 0.460 \\
  EigenWorms & 0.580 & 0.550 & 0.557 & 0.595 & 0.557 & 0.626 & 0.626 & 0.588 & 0.580 & 0.580 \\
  Epilepsy & 0.964 & 0.942 & 0.978 & 0.957 & 0.964 & 0.957 & 0.971 & 0.971 & 0.964 & 0.971 \\
  EthanolConcentration & 0.449 & 0.479 & 0.479 & 0.475 & 0.426 & 0.616 & 0.620 & 0.631 & 0.601 & 0.643 \\
  ERing & 0.956 & 0.959 & 0.952 & 0.956 & 0.956 & 0.952 & 0.948 & 0.948 & 0.944 & 0.956 \\
  FaceDetection & 0.564 & 0.570 & 0.562 & 0.564 & 0.576 & 0.575 & 0.573 & 0.567 & 0.574 & 0.568 \\
  FingerMovements & 0.510 & 0.490 & 0.520 & 0.540 & 0.530 & 0.580 & 0.590 & 0.590 & 0.570 & 0.550 \\
  HandMovementDirection & 0.405 & 0.432 & 0.432 & 0.432 & 0.446 & 0.446 & 0.392 & 0.419 & 0.432 & 0.486 \\
  Handwriting & 0.487 & 0.500 & 0.496 & 0.492 & 0.489 & 0.474 & 0.485 & 0.476 & 0.481 & 0.487 \\
  Heartbeat & 0.722 & 0.741 & 0.722 & 0.732 & 0.737 & 0.707 & 0.722 & 0.741 & 0.732 & 0.722 \\
  InsectWingbeat & 0.262 & 0.425 & 0.466 & 0.466 & 0.290 & 0.262 & 0.425 & 0.466 & 0.466 & 0.290 \\
  JapaneseVowels & 0.892 & 0.895 & 0.886 & 0.886 & 0.895 & 0.905 & 0.897 & 0.900 & 0.908 & 0.889 \\
  Libras & 0.872 & 0.872 & 0.872 & 0.872 & 0.872 & 0.889 & 0.878 & 0.883 & 0.894 & 0.883 \\
  LSST & 0.607 & 0.600   & 0.601 & 0.603 & 0.599 & 0.593 & 0.603 & 0.595 & 0.607 & 0.601 \\
  MotorImagery & 0.510 & 0.520 & 0.640 & 0.630 & 0.610 & 0.620 & 0.590 & 0.600 & 0.590 & 0.590 \\
  NATOPS & 0.817 & 0.839 & 0.833 & 0.822 & 0.817 & 0.856 & 0.811 & 0.844 & 0.828 & 0.828 \\
  PenDigits & 0.984 & 0.985 & 0.985 & 0.985 & 0.986 & 0.984 & 0.985 & 0.986 & 0.986 & 0.985 \\
  PEMS-SF & 0.809 & 0.798 & 0.798 & 0.803 & 0.798 & 0.803 & 0.809 & 0.809 & 0.809 & 0.809 \\
  PhonemeSpectra & 0.241 & 0.236 & 0.245 & 0.236 & 0.245 & 0.233 & 0.222 & 0.228 & 0.232 & 0.225 \\
  RacketSports & 0.855 & 0.862 & 0.855 & 0.882 & 0.855 & 0.895 & 0.888 & 0.901 & 0.901 & 0.888 \\
  SelfRegulationSCP1 & 0.782 & 0.785 & 0.785 & 0.785 & 0.782 & 0.819 & 0.816 & 0.812 & 0.816 & 0.795 \\
  SelfRegulationSCP2 & 0.467 & 0.539 & 0.472 & 0.472 & 0.456 & 0.506 & 0.506 & 0.517 & 0.500 & 0.483 \\
  SpokenArabicDigits & 0.980 & 0.983 & 0.981 & 0.977 & 0.978 & 0.984 & 0.983 & 0.980 & 0.985 & 0.985 \\
  StandWalkJump & 0.533 & 0.533 & 0.533 & 0.533 & 0.467 & 0.400 & 0.533 & 0.533 & 0.533 & 0.533 \\
  UWaveGestureLibrary & 0.912 & 0.916 & 0.912 & 0.916 & 0.916 & 0.912 & 0.912 & 0.916 & 0.906 & 0.916 \\
  \midrule
  \textbf{Average Effect Size} & -0.368 & -0.132 & -0.239 & -0.208 & -0.256 & 0.091 & 0.113 & 0.203 & 0.443 & 0.352 \\
  \bottomrule
  \end{tabular}}
  \label{tab:performance_metrics}
\end{table*}

Transforms show non-additive effects. RP and the DFT show higher performance. RP contributes more independently to performance improvements. The insignificance of DFT in the LOO study suggests that it may be represented by or diminishing other jointly aligned transforms. CWT with CGau2 shows less significance but higher effects in the LOO study.

\begin{researchq}
Sensitivity analysis for $\alpha$, $\beta$ and $\gamma$ in the trainin objective.
\end{researchq}

According to \textbf{Tab.}~\ref{tab:alpha_sen} and \textbf{Tab.}~\ref{tab:beta_sen}, the accuracy demonstrates moderate sensitivity to parameter variations, with optimal $alpha$=10.0 for SelfRegulationSCP2 (0.544) and $alpha$=0.1 for Heartbeat (0.732).  For $\beta=\gamma$ parameters, both datasets achieve peak performance at $\beta=\gamma$=0.1/10.0 (Heartbeat: 0.732) and $\beta=\gamma$=10.0/25.0 (SelfRegulationSCP2: 0.544), suggesting parameter robustness in mid-to-high ranges, which encourages larger punishment of forbidding orthogonality of the representations.

\begin{table}[htbp]\label{tab:alpha_sen}
  \centering
  \caption{Accuracy comparison with different $\alpha$ values.}
  \label{tab:results}
  \resizebox{\linewidth}{!}{\begin{tabular}{l *{5}{c}}
  \toprule
  Dataset & $\alpha=0.0$ & $\alpha=0.1$ & $\alpha=1.0$ & $\alpha=10.0$ & $\alpha=25.0$ \\
  \midrule
  SelfRegulationSCP2     & 0.528 & 0.528 & 0.528 & 0.544 & 0.533 \\
  Heartbeat & 0.712 & 0.732 & 0.712 & 0.722 & 0.712 \\
  \bottomrule
  \end{tabular}}
\end{table}

\begin{table}[htbp]\label{tab:beta_sen}
  \centering
  \caption{Accuracy comparison with different $\beta=\gamma$ values.}
  \label{tab:beta_results}
  \resizebox{\linewidth}{!}{\begin{tabular}{l *{5}{c}}
  \toprule
  Dataset & $\beta=\gamma=0.0$ & $\beta=\gamma=0.1$ & $\beta=\gamma=1.0$ & $\beta=\gamma=10.0$ & $\beta=\gamma=25.0$ \\
  \midrule
  SelfRegulationSCP2       & 0.539 & 0.539 & 0.528 & 0.544 & 0.544 \\
  Heartbeat  & 0.717 & 0.732 & 0.722 & 0.732 & 0.722 \\
  \bottomrule
  \end{tabular}}
\end{table}

\section{Experiment Details}\label{sec:exp_detail}

In this section, we exhibit the details of implementations of downstream tasks and baselines, characteristics of the datasets, evaluation metrics, and implementation details of our proposed methods, which are sufficient support for our demonstration of the experiments in \textbf{Sec.}~\ref{sec:empi}.

\subsection{Experimental Settings}\label{sec:exp_sett}

\begin{table}[h]
  \centering
  \caption{Statistics on the 30 UEA datasets are employed for classification assessment, while 12 specific subsets (denoted by $^*$) undergo clustering evaluation as per the methodology outlined in ~\cite{zhang2022multiview}.}
  \resizebox{\linewidth}{!}{\begin{tabular}{lccccc}
\toprule
Dataset	& \# Train &	\# Test &	\# Dim &	Length &	\# Class \\
\midrule

ArticularyWordRecognition$^*$ & 275 & 300 & 9 & 144 & 25 \\

AtrialFibrillation$^*$ & 15 & 15 & 2 & 640 & 3 \\

BasicMotions$^*$ & 40 & 40 & 6 & 100 & 4 \\

CharacterTrajectories & 1422 & 1436 & 3 & 182 & 20 \\

Cricket & 108 & 72 & 6 & 1197 & 12 \\

DuckDuckGeese & 50 & 50 & 1345 & 270 & 5 \\

EigenWorms & 128 & 131 & 6 & 17984 & 5 \\

Epilepsy$^*$ & 137 & 138 & 3 & 206 & 4 \\

EthanolConcentration & 261 & 263 & 3 & 1751 & 4 \\

ERing$^*$ & 30 & 270 & 4 & 65 & 6 \\

FaceDetection & 5890 & 3524 & 144 & 62 & 2 \\

FingerMovements & 316 & 100 & 28 & 50 & 2 \\

HandMovementDirection$^*$ & 160 & 74 & 10 & 400 & \textit{4} \\

Handwriting & 150 & 850 & 3 & 152 & 26 \\

Heartbeat & 204 & 205 & 61 & 405 & 2 \\

InsectWingbeat & 30000 & 20000 & 200 & 30 & 10 \\

JapaneseVowels & 270 & 370 & 12 & 29 & 9 \\

Libras$^*$ & 180 & 180 & 2 & 45 & 15 \\

LSST & 2459 & 2466 & 6 & 36 & 14 \\

MotorImagery & 278 & 100 & 64 & 3000 & 2 \\

NATOPS$^*$ & 180 & 180 & 24 & 51 & 6 \\

PenDigits$^*$ & 7494 & 3498 & 2 & 8 & 10 \\

PEMS-SF$^*$ & 267 & 173 & 963 & 144 & 7 \\

Phoneme & 3315 & 3353 & 11 & 217 & 39 \\

RacketSports & 151 & 152 & 6 & 30 & 4 \\

SelfRegulationSCP1 & 268 & 293 & 6 & 896 & 2 \\

SelfRegulationSCP2 & 200 & 180 & 7 & 1152 & 2 \\

SpokenArabicDigits & 6599 & 2199 & 13 & 93 & 10 \\

StandWalkJump$^*$ & 12 & 15 & 4 & 2500 & 3 \\

UWaveGestureLibrary$^*$ & 120 & 320 & 3 & 315 & 8 \\

\bottomrule
  \end{tabular}}
  \label{tab:uea_statistics}
\end{table}

\begin{table}[h]
  \centering
  \caption{Statistics of evaluated anomaly detection datasets.}
  \resizebox{\linewidth}{!}{\begin{tabular}{lccccc}
\toprule
Dataset	&  \# Entity  & \# Dim & Train length & Test length & Anomaly ratio (\%) \\
\midrule

UCR & 250 & 1 & 5302449 & 12919799 & 0.38 \\
\bottomrule
  \end{tabular}}
  \label{tab:ad_statistics}
\end{table}

\subsubsection{Baselines} We use 21 baselines for comparison, which are divided into two groups:

\textit{\underline{URL methods.}} We compare our \textbf{MMFA} framework with 6-time series URL baselines, including TS2Vec~\cite{yue2022ts2vec}, T-Loss~\cite{franceschi2019unsupervised}, TNC~\cite{tonekaboni2021unsupervised}, TS-TCC~\cite{eldele2021time},  TST~\cite{zerveas2021transformer} and CSL~\cite{liang2023contrastive}. All URL competitors are evaluated similarly to \textbf{MMFA} for a fair comparison.

\textit{\underline{Methods tailored to specific tasks.}} We include DTWD~\cite{bagnall2018uea} (1-NN classifier with dynamic time warping) and supervised methods MLSTM-FCNs~\cite{karim2019multivariate} (RNNs), TapNet~\cite{zhang2020tapnet} (attentional prototypes), ShapeNet~\cite{li2021shapenet} (shapelets), OSCNN~\cite{tang2020omni}, and DSN~\cite{xiao2022dynamic} (CNNs). Ensemble methods~\cite{lines2018time} are excluded for fairness. Supervised methods leverage true labels for feature learning, analogous to URL data augmentation, ensuring fair comparison with \textbf{MMFA}.

\textit{\underline{Task-specific benchmarks}} include the DTWD baseline (1-nearest-neighbor classifier with DTW distance) and supervised methods spanning RNN-based, attention-based, shapelet-driven, and CNN architectures. Ensemble methods were excluded for fairness. While supervised approaches leverage true labels—analogous to URL's data augmentation—the comparison remains valid as both paradigms align label-informed feature learning.

We assess six sophisticated clustering benchmarks, including feature selection based Time2Feat~\cite{bonifati2022time2feat}, dimension-reduction-based MC2PCA~\cite{li2019multivariate} and TCK~\cite{mikalsen2018time}, distance-based m-kAVG+ED and m-kDBA~\cite{ozer2020discovering}, deep learning-based DeTSEC~\cite{ienco2020deep}, and shapelet-based MUSLA~\cite{zhang2022multiview}.

Since no documented anomaly detection evaluations in segment-level settings exist, we create two raw MTS baselines using Isolation Forest for fair comparisons, models operating at each timestamp (IF-p) or within each sliding window (denoted as IF-s).

\subsubsection{Evaluation Metrics.} Standard metrics are used to assess downstream task performance. Accuracy (Acc)~\cite{bagnall2018uea} is applied for classification tasks. Clustering outcomes are measured using Rand Index (RI) and Normalized Mutual Information (NMI)~\cite{zhang2022multiview, zhang2018salient}. Anomaly detection employs F1-score~\cite{li2021multivariate}.

We quantified performance improvements using Cohen’s d, computing average effect sizes by comparing our method’s gains over baselines across datasets. The metric aggregates normalized differences relative to the pooled standard deviation of all methods’ performances, demonstrating our algorithm’s consistent superiority.

\subsection{Implementation Details.}\label{sec:impi_detail} The \textbf{MMFA} framework is implemented using PyTorch 1.10.2, and all experiments run on a Ubuntu machine equipped with Tesla A100 GPUs. Data augmentation methods are implemented using tsaug~\cite{tsaug} with default parameters. Finally, we report the performance after 100 epochs of training.

The majority of \textbf{MMFA}'s hyperparameters are consistently assigned fixed values across all experiments, devoid of any hyperparameter optimization. The coefficients $\alpha$ and $\beta$ in Eq.~\ref{eq:loss} are both assigned a value of 25, while $\gamma$ is set to 1. The SGD optimizer is employed to train all feature encoders with a learning rate fixed at $10^{-4}$.

The classification baselines and task-specific clustering baselines' outcomes are extracted from the cited publications~\cite{bagnall2018uea, li2021shapenet, tang2020omni, xiao2022dynamic, yue2022ts2vec, zhang2022multiview}. Our reproduced results cover other aspects.

Specifically, we use a window size of 100 and a representation dimension of 32 for the anomaly detection dataset. In the meanwhile we follow the protocols introduced in \cite{Wu2020CurrentTS} to evaluate accuracies and F1 scores for the UCR anomaly datasets, except for using isolation forest to produce multiple detections in each sub-dataset to fit with applicable usage.

The specific datasets utilized for each task are outlined below. The original datasets' dimensionalities (channel numbers) can overwhelm 2D transforms (e.g., RP, CWT) and ResNet encoders due to limited computation resources. Therefore, we perform average pooling on transformed multi-channel 2D features, capping dataset channels at 64. Raw time series channels remain unchanged for input into the time series encoder.

\section{Proofs of Theories}\label{sec:pot}

\subsection{\textbf{Theorem}~\ref{the:seman_sim}: Equivalence Between Eigenvalues and Distance Reduction of Spectral Embeddings.}\label{proof:eq}

\begin{proof}
    
We begin by expanding the expected squared difference:

\begin{equation}
\begin{aligned}
& \mathbb{E}_{(x, x^\prime)\sim p_{\text{\textit{sim}}}} [(f(x) - f(x^\prime))^2] \\
& = 2\mathbb{E}_{x \sim p_{\text{\textit{data}}}^\prime}[f(x)^2] - 2\mathbb{E}_{(x, x^\prime)\sim p_{\text{\textit{sim}}}}[f(x)f(x^\prime)] \\
&   \\ & \text{Expand the second term.} \\ \\
& =  2\mathbb{E}_{x \sim p_{\text{\textit{data}}}^\prime}[f(x)^2] \\ 
& \mathbb{E}_{x \sim p^\prime_{\text{\textit{data}}}, x^\prime \sim p_{\text{\textit{data}}}^\prime}[f(x) \int_{x^\prime} \frac { \frac 1 k p_{T_{(x)}, T_{(x^\prime)}}(x, x^\prime)} {p_{\text{\textit{data}}}^\prime(x)} f(x^\prime) dx^\prime] \\
&   \\ & \text{According to the eigenfunction property.} \\ \\
& = 2\mathbb{E}{x \sim p_{\text{\textit{data}}}^\prime}[f(x)^2] - \underbrace{2\mathbb{E}{x \sim p_{\text{\textit{data}}}^\prime}[(1-\lambda) f(x)^2]}_{\text{According to Eq~\ref{eq:eigen}.}} \\
& = 2\lambda\mathbb{E}_{x \sim p_{\text{\textit{data}}}^\prime}[f(x)^2]
\end{aligned}
\end{equation}

Thus, we have established the desired relationship:

\begin{equation}
\begin{aligned}
\mathbb{E}_{(x, x^\prime)\sim p_{\text{\textit{sim}}}} [(f(x) - f(x^\prime))^2]
& = 2\lambda\mathbb{E}{x \sim p_{\text{\textit{data}}}^\prime}[f(x)^2]
\end{aligned}
\end{equation}

\end{proof}

\subsection{\textbf{Theorem}~\ref{the:inv_est}: Multi-modal Invariance Estimation.}\label{proof:inv}

\begin{proof}
The proof of this theorem follows from the established inequalities detailed in Inequation~\ref{eq:invprime_prof}.

\begin{equation}
\begin{aligned}\label{eq:invprime_prof}
        \mathcal{L}_{inv}^\prime(Z) & = 
         \frac{1}{Nk(k + 1)}\sum_{i=1}^N(2\sum_{m=1}^d||z^{(0)}_i-z^{(m)}_i||^2_2 \\
        & + \sum_{\substack{m=1, n=1 \\ m \neq n}}^d||z^{(m)}_i - z^{(n)}_i||^2_2) \\
        & \leq \frac{1}{Nk(k + 1)}\sum_{i=1}^N[2\sum_{m=1}^d||z^{(0)}_i-z^{(m)}_i||^2_2 \\
        & + \sum_{\substack{m=1, n=1 \\ m \neq n}}^d(\underbrace{||z^{(0)}_i - z^{(m)}_i||^2_2+||z^{(0)}_i - z^{(n)}_i||^2_2)}_{\text{\textit{Triangle inequation.}}}] \\
        & = \frac{2}{N(k+1)} \sum_{i=1}^N\sum_{m=1}^d ||z^{(0)}_i - z^{(m)}_i||^2_2
\end{aligned}
\end{equation}
\end{proof}

\subsection{\textbf{Theorem}~\ref{the:orth}: Orthogonality of Eigenfunction-Based Representations.}

\begin{proof}

When we have the $n$ dimensional matrice $(D-w)$ which has finite rank, $||(D-w) - \mathbb{L}|| \xrightarrow{n \to \infty} 0 $ in an approximation sense. Thus, $\mathbb{L}$ is a compact operator. Therefor $\mathbb{L}=\sum_{i=1}^\infty \lambda_i \mathbb{E}_{x\sim p_{\text{data}}^\prime}[\cdot f_i(x)] f_i(x)$. For all $i \in [d_z] $, we have $\lambda_i \neq 0 $,  $ \mathbb{E}_{x \sim p_{\text{\textit{data}}}^\prime}[f_i(x)^2] = 1$. We prove that a set of $d_z$ eigenfunctions exists that are orthogonal to each other. The operator is symmetric.

\begin{equation}\label{eq:symetric}
    \mathbb{E}_{x \sim p_{\text{\textit{data}}}^\prime}[\mathbb{L}(f_i)(x) \cdot f_j(x)] = \mathbb{E}_{x \sim p_{\text{\textit{data}}}^\prime}[f_i(x) \cdot \mathbb{L}(f_j)(x)]
\end{equation}

According to the definition of eigenfunction, we have the equation below.

\begin{equation}\label{eq18}
    \lambda_i \mathbb{E}_{x \sim p_{\text{\textit{data}}}^\prime}[f_i(x) \cdot f_j(x)] =  \lambda_j \mathbb{E}_{x \sim p_{\text{\textit{data}}}^\prime}[f_i(x) \cdot f_j(x)]
\end{equation}

According to Eq.~\ref{eq18}, for any $\lambda_i \neq \lambda_j$ we have:
\begin{equation}
\begin{aligned}
     &   \forall i\neq j, \mathbb{E}_{x \sim p_{\text{\textit{data}}}^\prime}[f_i(x) \cdot f_j(x)] = 0
\end{aligned}
\end{equation}

When we have:

\begin{equation}
    \mathcal{S}_{\lambda} = \{f | \mathbb{L}(f)=\lambda f\} \ \ \text{dim}(\mathcal{S}_{\lambda}) \geq 2
\end{equation}

This generates a linear space.

\begin{equation}
\begin{aligned}
    & \mathbb{L}(ah + bg) = \mathbb{L}(ah) + \mathbb{L}(bg) = \lambda(ah + bg)  \\
    & \ \ \ \ \ \ \  a,b\in \mathbb{R} \ \ h,g \in \mathcal{S}_{\lambda} 
\end{aligned}
\end{equation}

We have $g_1 \in \mathcal{S}_\lambda$. Then, we remove the subspace spanned by $g_1$. Now we consider the space $\mathcal{S}^{\text{dim}(\mathcal{S}_{\lambda}) - 1} = \text{span}(\{g_1\})^\perp \cap \mathcal{S}_{\lambda}$. Then we have $g_i \in \mathcal{S}^{\text{dim}(\mathcal{S}_{\lambda}) - i + 1}$ where $\mathcal{S}^{\text{dim}(\mathcal{S}_{\lambda}) - i + 1}= \text{span}(\{g_j\}_{i-1})^\perp\cap \mathcal{S}_{\lambda}$. Finally, we have all the $\text{dim}(\mathcal{S}_{\lambda})$ orthogonal eigenfunctions with their eigenvalues equal to $\lambda$.

\end{proof}

\subsection{Spectral Selection Mechanism of Linearly Degenerated \textbf{MMFA} (LD-MMFA)}

This chapter establishes three foundational theorems for \textbf{MMFA} under linear degeneracy constraints. We introduce the gradually deduced closed-form solutions for the simple to complex construction of the training objectives. 

\textbf{Introduction of the Unsupervised Objectives.}\label{sec:recover} In linearly degenerated \textbf{MMFA}, the core objective is to learn a linear projection \( Z = XW \) that maps standardized high-dimensional data \( X \in \mathbb{R}^{n \times d} \) into a low-dimensional latent space \( Z \in \mathbb{R}^{n \times k} \), while preserving critical statistical structures across multiple transformed feature distributions.

The optimization problem is designed jointly.
\begin{itemize}
  \item (1) \textbf{Minimize Projection Complexity}: Penalize the Frobenius norm \( \|W\|_F^2 \) to select most significant spectrums to enhance generalizability. 
  \item (2) \textbf{Align Transformed Features}: Reduce discrepancies \( \|XW - F_i W_i'\|_F^2 \) between \( Z \) and each transformed view \( F_i W_i' \).  
  \item (3) \textbf{Enforce Orthogonality}: Constrain \( Z^TZ= W^\top X^\top X W = I_d \) and \( W_i'^\top F_i^\top F_i W_i' = I_d \) to ensure non-redundant, unit-covariance latent features.  
\end{itemize}

We compose these objectives and constraints, deduce their close-form solutions, and find out each spectral selection mechanism below.

\begin{lemma}\label{lem:PCA}LD-MMFA Without Transform Alignment Recovers PCA.

We use \( W^\top X^\top X W = I_d \) to forces orthogonality, but minimizing \( \|W\|_F^2 \) penalizes scaling inversely to variance. Thus, directions with maximal variance (largest \( \lambda_i \)) dominate, as they require minimal scaling to satisfy the orthogonality constraint.

The learning objective of LD-MMFA without transform alignment can be written as:

\begin{equation}
    \min_{W} \mathcal{L} = \|W\|_F^2 \quad \text{s.t.} \quad W^\top X^\top X W = I_d.
\end{equation}

Closed-form solution:  

\begin{equation}    
  W_{\text{opt}} = V_d \Lambda_d^{-1/2}, \quad \min_{W} \mathcal{L} = \sum_{i = 1}^d \lambda_{i}^{-1} 
\end{equation}

$\lambda_{X,i}$ is the $i$th largest singular value of $X^TX$.

\end{lemma}
\begin{proof}
  The constrained optimization problem is:
  \begin{equation}
      \min_{W} \|W\|_F^2 \quad \text{s.t.} \quad W^\top X^\top X W = I_d.
  \end{equation}
  
     Introduce symmetric Lagrange multiplier matrix \(\Lambda\):
     \[
     \mathcal{L} = \text{tr}(W^\top W) - \text{tr}\left(\Lambda^\top (W^\top X^\top X W - I_d)\right).
     \]
     Taking derivative w.r.t. \(W\) and setting to zero:
     \[
     W = X^\top X W \Lambda \quad \Rightarrow \quad X^\top X W = W (W^\top W)^{-1}.
     \]
  
     The equation \(X^\top X W = W (W^\top W)^{-1}\) implies \(W\)'s columns span an invariant subspace of \(X^\top X\). By the Spectral Theorem, this subspace must be spanned by eigenvectors of \(X^\top X\). Let \(X^\top X = V \Lambda V^\top\) be the eigenvalue decomposition with \(\Lambda = \text{diag}(\lambda_1, \dots, \lambda_p)\) (\(\lambda_1 \geq \cdots \geq \lambda_p\)).

     Express \(W = V_d Q\) where \(V_d\) contains the top \(d\) eigenvectors. The constraint becomes:
     \[
     Q^\top \Lambda_d Q = I_d \quad \Rightarrow \quad Q = \Lambda_d^{-1/2}.
     \]
     Substituting back, \(W_{\text{opt}} = V_d \Lambda_d^{-1/2}\). The objective becomes \(\sum_{i=1}^d \lambda_i^{-1}\), minimized by selecting the largest \(\lambda_i\) (smallest reciprocals).
  
  Thus, the solution recovers PCA's principal directions scaled by \(\lambda_i^{-1/2}\), proving equivalence.
  \end{proof}

\begin{lemma}\label{lem:CCA}LD-MMFA Selects Spectral Consensus with Single Transformed Feature.

The learning objective of LD-MMFA with single transform alignment can be written as:

\begin{equation}
\begin{aligned}
    \min_{W, W'} \mathcal{L} = \|W\|_F^2 + \|W'\|_F^2 + \|XW - FW'\|_F^2 \\
    \text{s.t.} \quad W^\top X^\top X W = W'^\top F^\top F W' = I_d.
\end{aligned}
\end{equation}  

Closed-Form Solution:  
\begin{equation}
\begin{aligned}
    & W_{\text{opt}} = V_X\Lambda_X^{-1/2} U_d, \quad W'_{\text{opt}} = V_F\Lambda_F^{-1/2} V_d, \\
    & \min_{W, W'} \mathcal{L} = \sum_{i=1}^{d} ( \frac{1}{\lambda_{X,i}} + \frac{1}{\lambda_{F,i}} - 2 \sigma_i ) + 2d \\
\end{aligned}
\end{equation}

where \( U_d, V_d \) are the top-\(d \) singular vectors of \( C = V_X\Lambda_X^{-1/2} X^\top F V_F\Lambda_F^{-1/2} \). And $\lambda_{X,i}, \lambda_{F,i}$ and $\sigma_i$ are the $i$th largest singular value of $X^TX, F'^\top F'$ and $\bar{X}^T \bar{F}$.

This optimization problem recovers canonical correlation analysis (CCA)~\cite{thompson2000canonical} for LD-MMFA and deep/kernel canonical correlation analysis (KCCA)~\cite{fukumizu2007statistical} for \textbf{MMFA}. This balances regularization penalties while aligning the transformed features.  
\end{lemma}

\begin{proof}  
  
     Define whitened matrices \(\tilde{X} = X V_X\Lambda_X^{-1/2}\) and \(\tilde{F} = F V_F\Lambda_F^{-1/2}\) (satisfying \(\tilde{X}^\top \tilde{X} = I\), \(\tilde{F}^\top \tilde{F} = I\)). Substitute \(W = V_X\Lambda_X^{-1/2} Q\), \(W' = V_F\Lambda_F^{-1/2} Q'\) to reduce constraints to \(Q^\top Q = I_d\), \(Q'^\top Q' = I_d\).  
  
     The loss becomes:  
     \[
     \mathcal{L} = \text{tr}(Q^\top \Lambda_X^{-1} Q) + \text{tr}(Q'^\top \Lambda_F^{-1} Q') - 2 \text{tr}(Q^\top C Q') + \text{const.},
     \]  
     where \(C = \tilde{X}^\top \tilde{F}\) is the whitened cross-covariance matrix.  
  
     Maximize \(\text{tr}(Q^\top C Q')\) via SVD with \(C = U \Sigma V^\top\). Optimal \(Q = U_d\), \(Q' = V_d\) (top-\(d\) singular vectors).  
  
     Align \(W, W'\) with maximal cross-correlation directions:  
     \[
     W_{\text{opt}} = V_X\Lambda_X^{-1/2} U_d, \quad W'_{\text{opt}} = V_F\Lambda_F^{-1/2} V_d.
     \]  
     Minimal loss derived as \(\sum_{i=1}^d (\lambda_{X,i}^{-1} + \lambda_{F,i}^{-1} - 2\sigma_i) + 2d\), where \(\sigma_i\) are top-\(d\) singular values of \(C\).

\end{proof}
    
\begin{lemma}\label{lem:multiCCA}
Multiview Alignment via Joint Cross-Covariance Spectrum.
The concatenated matrix \( C \) encodes all pairwise correlations between \( X \) and each \( F_i \). Its SVD extracts a consensus subspace \( U_d\) in \( X \) that maximizes the total correlation with all \( F_i \)-views.

Optimization Problem:  
\begin{equation}
\begin{aligned}
\min_{W, \{W_i'\}} \mathcal{L} = \|W\|_F^2 + \sum_{i=1}^p \left( \|W_i'\|_F^2 + \|XW - F_i W_i'\|_F^2 \right) \\
\text{s.t.} \quad W^\top X^\top X W = W_i'^\top F_i^\top F_i W_i' = I_d.
\end{aligned}
\end{equation}  
Closed-Form Solution:  
\begin{equation}
\begin{aligned}
& W_{\text{opt}} = V_X\Lambda_X^{-1/2} U_d \quad W_{i,\text{opt}}' = V_{F_i}\Lambda_{F_i}^{-1/2} V_{i,d}, \\
& \min_{W, W'} \mathcal{L} = \sum_{j=1}^{d} ( \frac{1}{\lambda_{X,j}} + \sum_{i=0}^p \frac{1}{\lambda_{F,i,j}} - 2 \sigma_j ) + 2d \\
\end{aligned}
\end{equation}  

$\lambda_{X,j}, \lambda_{F,i,j}$ and $\sigma_j$ are the $j$th largest singular value of $X^TX, F_i'^\top F_i'$ and $\bar{X}^T \bar{F}$. where \( U_d \) and \( V_{i,d} \) are derived from the SVD of \( C = [\tilde{X}^\top \tilde{F}_1 \; \cdots \; \tilde{X}^\top \tilde{F}_p] \).

\end{lemma}

\begin{proof}

Extending \textbf{Lemma}~\ref{lem:CCA} to multiple views, we first whiten \( X \) and each \( F_i \):  
\begin{equation}
\begin{aligned}
\tilde{X} = X V_X\Lambda_X^{-1/2}, \quad \tilde{F}_i = F_i V_{F_i}\Lambda_{F_i}^{-1/2}.
\end{aligned}
\end{equation} 
Substituting \( W = V_X\Lambda_X^{-1/2} Q \) and \( W_i' = V_{F_i}\Lambda_{F_i}^{-1/2} Q_i' \), the constraints reduce to \( Q^\top Q = I_d \) and \( Q_i'^\top Q_i' = I_d \).  

The objective simplifies to maximizing \( \sum_{i=1}^p \text{tr}(Q^\top \tilde{X}^\top \tilde{F}_i Q_i') \), which aggregates cross-view correlations. This is equivalent to maximizing \( \text{tr}(Q^\top C [Q_1'^\top \; \cdots \; Q_p'^\top]^\top) \), where \( C = [\tilde{X}^\top \tilde{F}_1 \; \cdots \; \tilde{X}^\top \tilde{F}_p] \).  

Applying SVD to \( C \), we write \( C = U \Sigma [V_1^\top \; \cdots \; V_p^\top]^\top \), where \( U \) and each \( V_i \) are orthogonal. The optimal \( Q \) is the first \( d \) columns of \( U \), while each \( Q_i' \) corresponds to the first \( d \) columns of \( V_i \).  The minimal loss can be derived with the result.

\end{proof}

\subsection{Theorem~\ref{theorem:recover}: \textbf{MMFA}'s recovery of KPCA and KCCA.}\label{sec:recoverk}

The following proof demonstrates that under linear degeneracy constraints, \textbf{MMFA} recovers Kernel Principal Component Analysis (KPCA)~\cite{kpca} and Kernel Canonical Correlation Analysis (KCCA)~\cite{fukumizu2007statistical} when applied to kernel-mapped data. The linear solutions derived in \textbf{Lemmas} \ref{lem:PCA}–\ref{lem:multiCCA} generalize to their kernelized counterparts through the kernel trick, where input data \(X\) and transformed features \(F_i\) are implicitly mapped into reproducing kernel Hilbert spaces (RKHS).

\begin{proof}
When \(p = 0\) (no transformed features), we apply LD-MMFA to kernel-mapped data \(\Phi_X(X)W\) with \(p = 0\). According to \textbf{Lemma}~\ref{lem:PCA}, the solution to LD-MMFA reduces to KPCA,

When aligning a single transformed feature \(\Psi_F(F)W'\) (kernel-mapped from another view \(F\)), we apply LD-MMFA to kernel-mapped data \(\Phi_X(X)W\) and \(\Phi_F(F)W'\) with \(p = 1\). According to \textbf{Lemma}~\ref{lem:CCA} the solution recovers KCCA, selecting directions of maximal correlation between kernel spaces.

For multiple transformed features \(\{\Psi_{F,i}(F_i)W_{F,i}\}\), the solution aligns a consensus subspace across all kernel-induced views, according to \textbf{Lemma}~\ref{lem:multiCCA} generalizing multi-view KCCA.
\end{proof}

\section{Conclusion}\label{sec:conc}
  
In this study, we propose a framework aimed at mitigating the limitations arising from intrinsic feature engineering in Universal Representation Learning (URL) for Multivariate Time Series (MTS). The incorporation of feature alignment and regularization methods forms a revisitation of the spectral selection mechanism for constructing URL framework, surpassing existing state-of-the-art approaches.

Looking ahead, this study unveils several promising avenues. For instance, delving deeper into transform property deduction and more efficient learning methods could yield more comprehensive representations. The integration of broader external knowledge or domain-specific information holds the potential to further enrich these representations and accelerate the learning process.

\section*{Acknowledgments}
This work is supported by the National Natural Science Foundation of China (NSFC) (72371084)


\bibliography{example_paper}
\bibliographystyle{IEEEtran}

\newpage

 




\vfill

\end{document}